\documentclass[10pt,twocolumn,letterpaper]{article}

\usepackage{cvpr}              
\usepackage{comment}
\usepackage{float}
\usepackage[dvipsnames]{xcolor}

\definecolor{cvprblue}{rgb}{0.21,0.49,0.74}
\usepackage[pagebackref,breaklinks,colorlinks,citecolor=cvprblue]{hyperref}
\usepackage{graphicx}
\usepackage[normalem]{ulem}
\useunder{\uline}{\ul}{}
\usepackage[accsupp]{axessibility}
\def\paperID{10417} 
\def\confName{CVPR}
\def\confYear{2024}

\title{ChAda-ViT : Channel Adaptive Attention for Joint Representation Learning of Heterogeneous Microscopy Images}

\author{
    Nicolas Bourriez$^{1}$\footnotemark[1] \quad Ihab Bendidi$^{1,2}$\footnotemark[1] \quad Ethan Cohen$^{1,3}$\footnotemark[1] 
    \quad Gabriel Watkinson$^{1,3}$\\
     \quad Maxime Sanchez$^{1,3}$ \quad Guillaume Bollot$^{3}$
     \quad Auguste Genovesio$^{1}$ \vspace{0.3em} \\
    {\normalsize $^1$Ecole Normale Supérieure PSL, Paris, France} \\
    {\normalsize $^2$Minos Biosciences, Paris, France} \\
    {\normalsize $^3$Synsight, Evry, France} \vspace{0.3em} \\
    {\normalsize \texttt{firstname.lastname@ens.psl.eu}}\\
    {\normalsize $^*$Equal Contribution}
}

\begin{document}

\maketitle
\begin{abstract}
Unlike color photography images, which are consistently encoded into RGB channels, biological images encompass various modalities, where the type of microscopy and the meaning of each channel varies with each experiment. Importantly, the number of channels can range from one to a dozen and their correlation is often comparatively much lower than RGB, as each of them brings specific information content. This aspect is largely overlooked by methods designed out of the bioimage field, and current solutions mostly focus on intra-channel spatial attention, often ignoring the relationship between channels, yet crucial in most biological applications. Importantly, the variable channel type and count prevent the projection of several experiments to a unified representation for large scale pre-training. In this study, we propose ChAda-ViT, a novel Channel Adaptive Vision Transformer architecture employing an Inter-Channel Attention mechanism on images with an arbitrary number, order and type of channels. We also introduce IDRCell100k, a bioimage dataset with a rich set of 79 experiments covering 7 microscope modalities, with a multitude of channel types, and counts varying from 1 to 10 per experiment. Our architecture, trained in a self-supervised manner, outperforms existing approaches in several biologically relevant downstream tasks. Additionally, it can be used to bridge the gap for the first time between assays with different microscopes, channel numbers or types by embedding various image and experimental modalities into a unified biological image representation. The latter should facilitate interdisciplinary studies and pave the way for better adoption of deep learning in biological image-based analyses. 

\footnotetext[3]{Code, Data \& Model weights : \href{https://github.com/nicoboou/chadavit}{https://github.com/nicoboou/chadavit}}

\end{abstract}    
\section{Introduction}
\label{sec:intro}

\begin{figure}[tb!]
\begin{center}
\centerline{\includegraphics[width=\columnwidth]{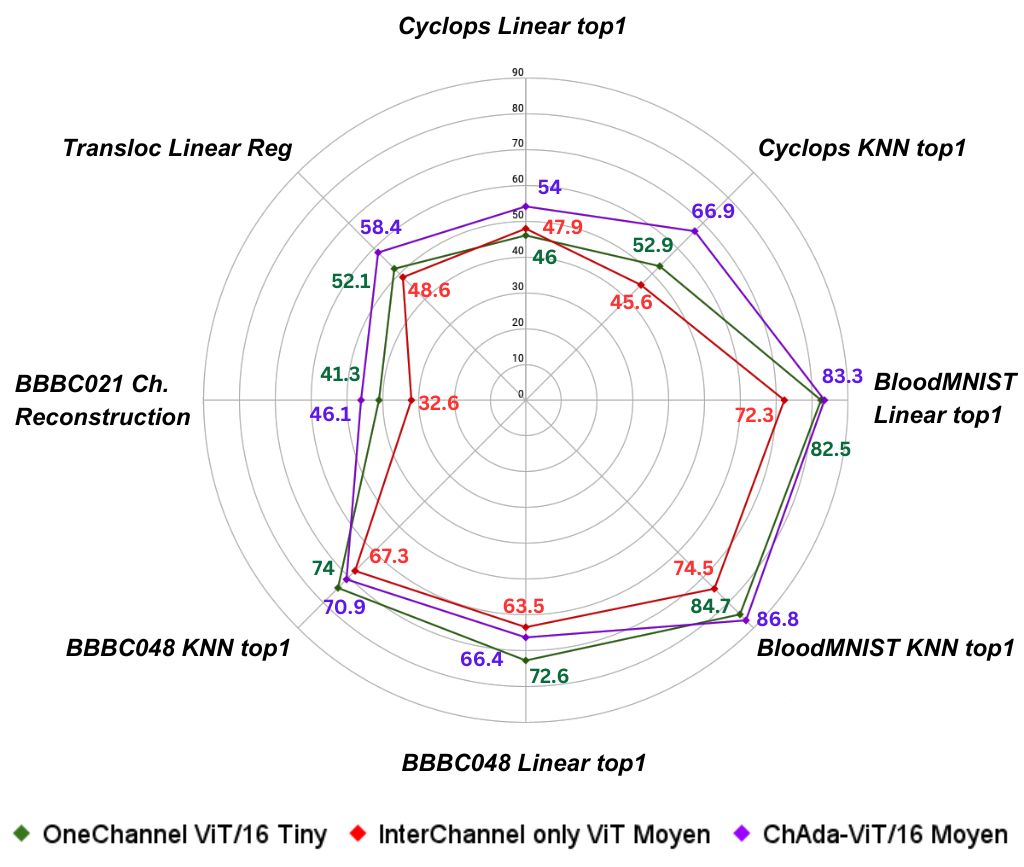}}
    \caption{Performance comparison on downstream tasks, showcasing ChAda-ViT's superiority in 6 out of 8 tasks compared to existing approaches\cite{microsnoop} using CLS token only. \(R^2\) scores, normalized to 0-100, are presented for BBBC021 Channel Reconstruction and Nuclear Translocation prediction tasks. This success is attributed to the combined use of Intra-Channel and Inter-Channel Attention. Evaluation on all tokens is detailled in Appendix.}
    \label{fig:main_results}
\end{center}
\vskip -0.3in
\end{figure}

Revolutionizing the field of image processing, Convolutional Neural Networks (CNNs) set the stage for unprecedented advancements \cite{LeCun1989BackpropagationAT,Krizhevsky2012ImageNetCW,He2015DeepRL}. Meanwhile, originally designed for Natural Language Processing (NLP), transformers emerged \cite{Vaswani2017AttentionIA} later in computer vision as Vision Transformers (ViTs) \cite{Dosovitskiy2020AnII}. ViTs excel in handling large datasets, detecting long-range dependencies \cite{Rao2021DynamicViTEV} and capturing spatial correlations, often surpassing the capabilities of their CNN counterparts \cite{Carion2020EndtoEndOD,Dai2020UPDETRUP,touvron2020training,Yuan2021TokenstoTokenVT,Meng2021AdaViTAV,Xie2021SegFormerSA}. The adoption of ViTs has spurred important applications in many tasks\cite{Yu2021MILVTMI,Chen2021TransUNetTM,Sha2022MITformerAM,CrossZamirski2021LabelfreePO}, as well as the creation of foundation models that have significantly improved performance across a variety of domains\cite{oquab2023dinov2,Filiot2023scalingwithMIM}, marking a new era of versatility and robustness in the application of these transformative technologies.

Bioimaging is however distinct, marked by its sparsity and lack of standardization, a contrast to the rapid advancements seen in general computer vision. Unlike conventional pictures in RGB color format, microscopy images span a plethora of specialized types of image across various channels \cite{Goltsev2017DeepPO,vonChamier2021DemocratisingDL,Hickey2021SpatialMO}. Each channel, marked by a different staining or microscopy imaging technique, discloses unique biological information that, in many case, is specific to the very assay being imaged. This multimodality of images in term of channel numbers and types is pivotal as it underpins the heterogeneity inherent in bioimaging data. The capability to discern and leverage inter-channel relationships is paramount in this case as it necessarily lead to a richer capture of biological phenomena, thereby holding significant promise in advancing biological observation. \cite{Chandrasekaran2022ThreeMI}.

However, due to the prominent use of RGB pictures in computer vision literature, transfer of current methodologies to biological images predominantly advocate for individual channel encoding \cite{Doron2023UnbiasedSM,microsnoop,kim2023selfsupervision}. A stance that, albeit practical for certain tasks, overlooks the potential insights harbored in the interplay between channels and the information they represent. Importantly, this approach thwarts the re-usability of pre-trained models across diverse studies\cite{Cohen2022CellPT,Cohen2021AuraNetRS,Harkness2022LeveragingAT,Masud2022ComparisonOS}. Models tailored to a specific microscopy configuration may yield compromised performance due to data scarcity, possibly driving spurious correlations over valuable biological features \cite{Schlkopf2021TowardCR}. These approaches also overlook the opportunity to exploit the vast heterogeneous biological data available \cite{Hartley2021TheBA,Williams2017}. A unified architecture accommodating the diverse nature of bioimaging data would not only facilitates the establishment of a common biological embedding space for various vision tasks but also heralds the potential of crafting a single pre-trained model. Such a model could serve as a linchpin for broader studies across different biological tasks, enabling comparative analyses, and studying correlations in a streamlined and unified analytical space. This consolidation could significantly accelerate and streamline analysis, fostering a quicker adoption of deep learning within the biological community for image-based studies.

Through attempting to resolve these issues, the main contributions of this paper are threefold:

\begin{itemize}
\item The introduction of a heterogeneous bioimage dataset encompassing various channel types and numbers, as well as a variety of microscopy imaging techniques used to acquire these channels.

\item The introduction of a backbone architecture of Vision Transformers capable of handling bioimage datasets with different numbers and types of channels through a masking strategy coupled with intra and inter-channel attention, while achieving state-of-the-art results in a number of biologically relevant tasks compared to the usual ViT based approach for biological images.

\item For the first time, to the best of our knowledge, we present a unified embedding space for any microscopy image dataset, bridging the gap between different heterogeneous datasets and opening the door to cross-modal imaging studies.

\end{itemize}
\section{Related Works}
\label{sec:related_works}

\textbf{Microscopy Image analysis. }Advancements in image processing for biological applications have significantly contributed to high-throughput assays analyses and functional genomics. Open tools such as CellProfiler \cite{mcquin2018cellprofiler} have been integral, facilitating cell analysis with simple and efficient approaches, as well as modular image analysis pipelines for 3D image stacks and cloud-based processing to handle the surge in \textit{biological big data}. The introduction of CNNs has further augmented bio-imaging, providing rapid and efficient solutions to tasks such as phase unwrapping, subtle phenotype analysis and multi-parametric cell classification and analysis. \cite{bourou2023unpaired,lamiable2023revealing,frontierscnn, cnn_lensfree}. Vision Transformers have leveraged their ability to effectively learn long-range dependencies in biological data, presenting new opportunities and addressing remaining challenges in bioimage analysis \cite{li2023challenges}. Moreover, recent studies \cite{nofreelunch} highlight the significant influence of transformation design on feature learning in microscopy images, an aspect that underscores the need for biology-specific considerations in self-supervised learning (SSL). Additionally, SSL methods employing vision transformers, such as DINO \cite{dino}, have outshined traditional tools and achieved superior performance in numerous biological tasks, offering better classification of chemical perturbations and clustering gene families \cite{kim2023selfsupervision}, and advancing morphological profiling, with enhanced capabilities in encoding complex cellular morphology without manual supervision \cite{Doron2023UnbiasedSM}. These developments mark a shift toward more automated and sophisticated frameworks in bioimage analysis, crucial for navigating the increasing complexity and volume of biological data.

\textbf{Unified Microscopy Image Representation. }In computational biology, achieving a unified representation space suitable for diverse microscopy techniques remains an ongoing challenge. While cell painting methods \cite{cell_painting} offer a data-rich environment for in-depth analysis within their specific domain \cite{zheng2022cross,Moshkov2023-jg,gabriel2023weakly}, the scarcity of data in other experimental types creates a bottleneck for wider application. Approaches like Microsnoop \cite{microsnoop} address this by operating in a one channel encoding regime, encoding each channel independently and subsequently concatenating these into a final representation. However, this approach leads to variable-sized embeddings depending on channel configurations, which impedes the integration of data for cross-experiment studies. CytoImageNet's strategy \cite{cytoimagenet} to average channel information into a single-channel dataset resolves the representation space inconsistency but at the cost of losing detailed multi-channel information. 
To the best of our knowledge, no method offers a joint representation inclusive of all types of microscopy and channel configurations.

\section{Dataset}
\label{sec:dataset}

We introduce the IDRCell100K image dataset, a collection of biological images, purposefully curated from the extensive and varied Image Data Resource platform \cite{Williams2017}. This section outlines the process for selecting and refining these images and provide details on these biological assays, with the explicit goal of encompassing a heterogeneous distribution of data. Our selection, based on metadata provided with these experiments, covered various microscopy techniques to encapsulate a diverse array of imaging modalities, ensuring the dataset's breadth in representing biological information. Efforts were made to minimize experimental and imaging biases, striving for a balanced representation up to a feasible extent, thereby reducing dependency on each image modality or experiment. Further details on the equitable distribution of images across different microscopy modalities within the dataset are available in the Appendix.

\textbf{Data Source Heterogeneity. }To create a well-rounded dataset, we focused on cell culture experiments from the Image Data Resource. We picked 79 distinct experiments conducted under different conditions and for different scientific purposes. These experiments employed 7 types of microscopy techniques and fell into 6 categories of study. 

\textbf{Data Selection. }As the number of images differ from one experiment to the other, we carefully chose 1,300 images from each selected experiment, in order to keep the final dataset balanced. These images come from experiments using different methods and include a wide range of channels monitoring for various components of the cells. Altogether, we end up with 308,898 single channel images, which we resized to 224x224 pixels from a variety of original sizes. When combined, it resulted to 104,093 multiplexed microscopy images containing cells at various scales, with each image made from one to up to 10 different channels.

\textbf{Implementation details. }Due to the lack of a dedicated Application Platform Interface (API), the retrieval of these images was performed through automated authorized webscraping of the Image Data Ressource platform. This process was performed on a distributed High-Performance-Computing CPUs cluster, using HTCondor cluster manager software \cite{condor-practice} with 10 processes per node. In this settings, creation of the IDRCell100k Dataset took two weeks.

\section{Methodology}
\label{sec:proposed_method}

\subsection{Problem Formulation}
\label{sec:problem_formulation}

\begin{figure*}[tb!]
\begin{center}
\centerline{\includegraphics[width=\textwidth]{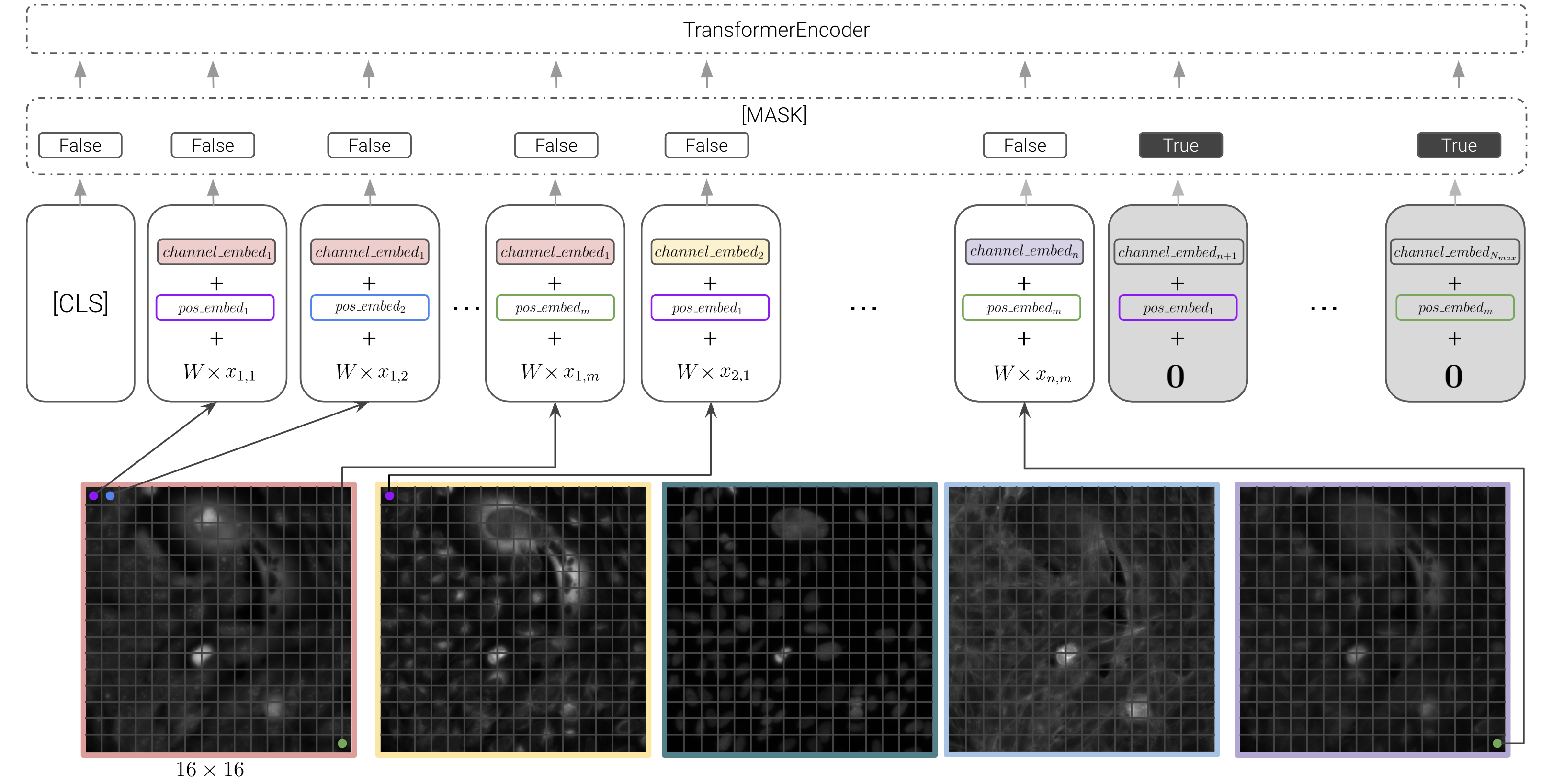}}
    \caption{The ChAda-ViT model architecture, displaying the proposed channel-adaptive embedding process. This figure illustrates the token padding and masking approach for image data with an arbitrary number of channels, the split of channels into patches, and the integration of positional and channel-specific embeddings to reach a fixed size input. We use the same positional emedding for patches in the same position accross channels, and the same channel embedding for all patches of each channel.}
    \label{fig:ChAda_vit}
\end{center}
\vskip -0.3in
\end{figure*}

Let \(I\) denote a single image from the dataset, extracted from a union of spaces \(\bigcup \mathbb{R}^{H \times W \times n_i}\), where \(n_i\) represents the number of channels in image \(I_i\), and \(1 \leq n_i \leq N_{\text{max}}\), with \(N_{\text{max}}\) being the maximum number of channels across all images.

The objective is to find a projection function \(\Phi\) that maps an image \(I_i\) to a latent space \(\mathbb{R}^{K}\), formalized as:
\begin{equation}
\Phi: \bigcup \mathbb{R}^{H \times W \times n_i} \rightarrow \mathbb{R}^{K}
\label{eq:projection_function}
\end{equation}
\begin{equation}
l_i = \Phi(I_i)
\label{eq:latent_representation}
\end{equation}
where \(K\) is the dimensionality of the latent space for the image \(I_i\).

\subsection{Gold Standard Approach}
\label{sec:baseline}

Existing works in literature \cite{cytoimagenet,microsnoop}  such as Microsnoop adopt a One Channel Encoding approach, which results for Image \(I_i\) with \(n_i\) channels into a generalized function mapping \(\Phi'\) defined as :

\begin{equation}
\Phi': \bigcup \mathbb{R}^{H \times W \times n_i} \rightarrow \bigcup \mathbb{R}^{K' \times n_i}
\label{eq:projection_function_prime}
\end{equation}
where \( K' \) is the dimensionality of the latent space for a single channel.

This overarching function is effectively realized by processing each channel \( j \) of an image \( I_i \) independently. A function \(\Psi\) is trained to project these individual channels into a latent space  \( \mathbb{R}^{K'} \) :

\begin{equation}
\Psi: \mathbb{R}^{H \times W} \rightarrow \mathbb{R}^{K'}
\label{eq:psi_function}
\end{equation}
\begin{equation}
l_{ij} = \Psi(I_{ij})
\label{eq:latent_representation_psi}
\end{equation}
where \( l_{ij} \) represents the latent feature of the \( j \)-th channel of the \( i \)-th image.

For images with disparate channel counts \( n_i \), the individual latent features \( l_{ij} \) are concatenated to form a representation \( l_i \) for each image:
\begin{equation}
l_{i} = \bigoplus_{j=1}^{n_i} l_{ij}
\label{eq:aggregated_representation}
\end{equation}
The dimensionality of this representation \( l_i \) aligns with the shape \( K' \times n_i \). Therefore it depends on the original number of channels \(n_i\) in the image ( \( l_{i} \in \bigcup \mathbb{R}^{K' \times n_i} \)) which makes it impossible to train a single such a model on a large heterogeneous bioimage dataset as the one we assembled. Also it does not provide a way to encode various dataset into a single representation.

\subsection{Our Proposed Approach}
\label{sec:our_method}

\begin{table*}[tb]
\centering
\resizebox{\textwidth}{!}{%
\begin{tabular}{|c|c|c|c|c|c|c|}
\hline
\textbf{Dataset} & \textbf{Downstream Task(s)} & \textbf{Granularity} & \textbf{Biological Application} & \textbf{Dataset size} & \textbf{Shape} & \textbf{Metric} \\ \hline
{BloodMNIST\cite{Yang2023}}             & Clustering + Classification & Single-Cell & Cellular types                     & 17092 & 28*28*3               & Accuracy \\ \hline
{CyclOPS\footnotemark[2]}                & Clustering + Classification & Single-Cell & Protein Localization labelling & 28166 & 64*64*2               & Accuracy \\ \hline
{BBBC048\footnotemark[3]}                & Clustering + Classification & Single-Cell & Cell Cycle Stages              & 32266 & 66*66*3               & Accuracy \\ \hline
{NF-kB Nuclear Transloc\cite{lamiable2023revealing}} & Regression                  & Single-cell & Nuclear Translocation          & 1000 & 256*256*3 & R2       \\ \hline
{BBBC021\cite{bbbc021}}                & Generation                  & Whole Slide & Imaging                        & 13200 & 1024*1280*3           & R2,MSE,MAE       \\ \hline
\end{tabular}%
}
\caption{Overview of biological datasets used to compare the ChAda-ViT and One Channel ViT models across clustering, classification, regression, and generative tasks, highlighting the diversity in biological applications, dataset sizes, and complexity. Performance metrics include Top1 accuracy, R2, MSE and MAE, corresponding to the task-specific objectives.}
\label{tab:datasets}
\vskip -0.15in
\end{table*}

We introduce ChAda-ViT, the \textbf{Ch}annel \textbf{Ada}ptive \textbf{Vi}sion \textbf{T}ransformer, a unified architecture for a model \(\Phi\) as defined in Eq. \ref{eq:projection_function}, capable of encoding images with heterogeneous channel dimensions \(n_i\) into a single fixed size embedding space, through the introduction of Inter-Channel Attention, on top of the regular intra-channel or spatial attention. This is a consequence of leveraging the principles of token padding and masking -- a technique originally established in NLP transformers \cite{Vaswani2017AttentionIA} and partially adapted to Self Supervised learning in ViTs \cite{he2021masked, assran2023selfsupervised} --, and introducing the concept of channel embeddings, as shown in Figure \ref{fig:ChAda_vit}, to accommodate the variable number of channels present in different images.

The proposed approach patchifies each channel separately instead of considering them altogether. Each channel \( j \) of an image \( I \) with dimensions \( H \times W \times n \) is split into non-overlapping patches \( P_{j, x, y} \). Each patch at spatial location \( (x, y) \) and channel \(j\) is of size \( p \times p \), where \( p = 16 \) is the preferred ViT configuration. These patches are then projected into a lower dimension with a shared 2D convolutional layer.

To standardize the input of the Transformer, we employ a padding strategy that compensates for images \( I \) with fewer channels than \( N_{\text{max}} \), the maximum number of channel over the dataset. Padding tokens extend the patch sequence of each image to match the length of \( N_{\text{max}} \times m \), where \( m \) is the number of patches per channel. Thus, the padded sequence, \( \text{Seq}_{\text{pad}}(I) \), ensures any image is transformed into a fixed vector size to feed the the model.
To maintain the integrity of the self-attention mechanism within the Transformer, we apply a binary masking strategy during attention computation. A mask is created for each image in the batch, marking the locations of padding tokens to ensure these are excluded from contributing to the self-attention mechanism. This method allows ChAda-ViT to focus solely on the meaningful data patches and preserve the inter-channel and intra-channel attention accuracy.

We also introduce the concept of channel embeddings, which focus on preserving channel information. Each patch \( P_{j, x, y} \) is enriched with both \textit{positional} and \textit{channel-specific} embeddings to preserve its spatial context and channel identity. Positional embeddings  \( pos_{x,y}  \)  ensure spatial information is maintained across all channels, while channel embeddings \( chan_j \) mark each patch with its respective channel origin. The dual embedding strategy allows the model to distinguish between patches of different channels located at the same spatial position. Both types of embeddings are learnable parameters, fine-tuned during the training process to optimize the representation of spatial and channel information within the unified embedding space.

\subsection{Model Architecture}

We use Vision Transformer (ViT) models to examine whether incorporating Inter-Channel attention -- achieved by channel-specific patchification and token padding and masking -- in addition to Intra-Channel attention enhances model performance on biological image tasks. We employ a ViT-Tiny architecture as the backbone. The model, employs a shared 2D convolutional layer to embed each token with an embedding dimension of 192. Due to the dataset's maximum channel count (\(C_{max}\)) being 10 and the image size being 224x224, the ChAda-ViT model processes a significantly high number of input tokens--10 times more than a standard ViT-Tiny and 50\% more than a ViT-Large. To avoid confusion with traditional ViT size nomenclature, we adopt a distinct model name, dubbed ChAda-ViT Moyen (French for \textit{Average}), reflecting its expanded width. Experiments with  different ChAda-ViT architecture sizes (Grand and Petit) to confirm the scaling laws of our method are available in the Appendix.
The proposed approach is compared to Microsnoop One Channel approach \cite{microsnoop}, using  a standard ViT for a fair comparison, modelling the function \(\Psi\) as defined in Eq. \ref{eq:psi_function}, to serve as the baseline, using similar backbone, embedding size, and token per channel count to evaluate the effects of our channel-adaptive contribution. This baseline encodes each channel separately with Intra-Channel attention, and then combines the resulting \textit{CLS} token representations into a \(n_i \times 192\)-dimensional image representation. Furthermore, we introduce an Inter-Channel only ViT variant as an ablation study of the inter-channel attention only, where each channel is treated as a distinct single patch of size 224x224, as opposed to the 16x16 patch size used in the one-channel ViT and ChAda ViT. Each of these full-sized channel patches is tokenized into a 192-dimensional vector, compelling the model to focus its attention solely on the features derived from the relationships between the individual channels by eschewing Intra-Channel considerations.

\footnotetext[2]{The dataset can be accessed on Kaggle: \href{https://www.kaggle.com/datasets/stanleyhua/cyclops-protein-loc}{CYCLoPs Dataset}.}

\footnotetext[3]{More details on the dataset are found here : \href{https://bbbc.broadinstitute.org/BBBC048/}{BBBC048}.}
\section{Experiments}
\label{sec:experimental_setup}

\noindent \textbf{Model Training. }Given the heterogeneous nature of the data, with its assortment of unrelated experiments, diverse image types, channel configurations, cell lines, and labels, usage of standard supervised approaches presents a unique challenge due to the strong label variability and occasional label absence.. Therefore, we aim to obtain broad representations through self-supervised learning (SSL), assessing these models on specialized downstream tasks on biological images. The three models are thus trained on the IDRCell100k dataset we created with DINO \cite{dino} as the SSL strategy for 400 epochs. ChAda-ViT Moyen and Inter-Channel only ViT Moyen are set with a base learning rate of 0.0001, while the one-channel ViT-Tiny is trained with a base learning rate of 0.005 for the sake of stability during training. A batch size of 256 multiplexed images per GPU is maintained for both models. Training employs a cosine annealing scheduler to optimize the learning rate over time. The ChAda-ViT Moyen undergoes training on 32 A100 80GB GPUs distributed across four nodes, for a total training time of 2080 GPU hours. 

\begin{figure*}[tb!]
\begin{center}
\centerline{\includegraphics[width=\textwidth]{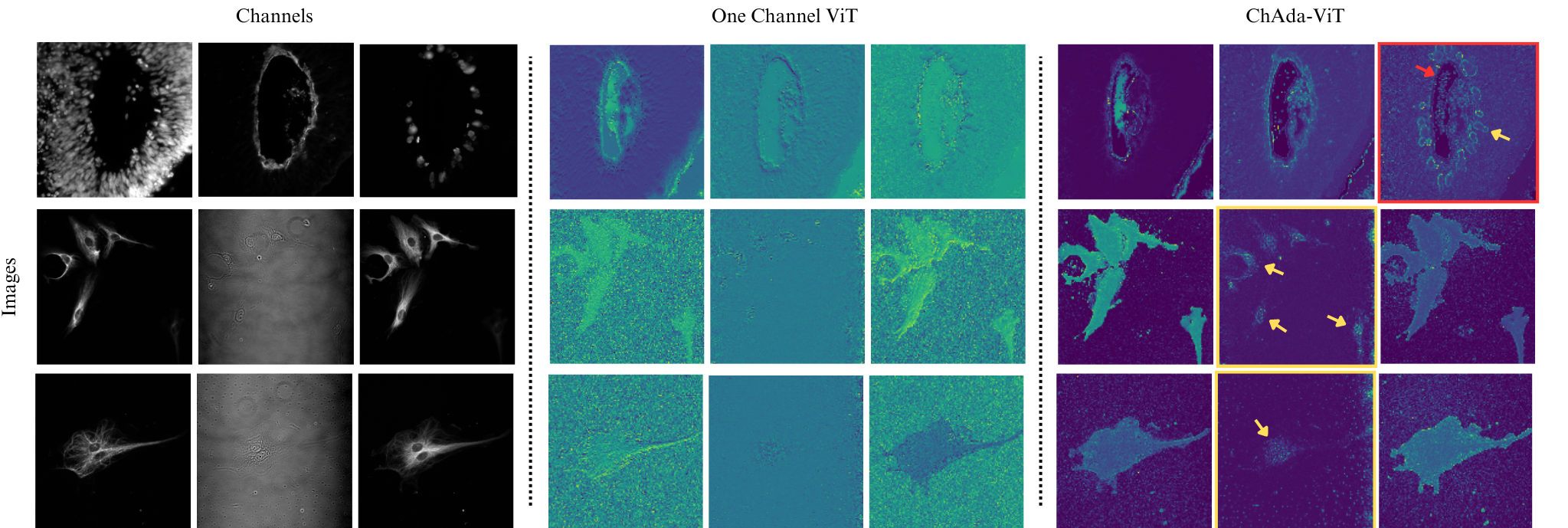}}
    \caption{Comparison of the last layer self-attention maps between One Channel ViT and ChAda-ViT on IDRCell100k image channels. ChAda-ViT, utilizing Inter-Channel Attention, effectively discerns significant cross-channel correlations (red arrow), focusing on spatially relevant areas in each channel (yellow arrows). This mechanism enables ChAda-ViT to identify critical biological features that might be overlooked with a single-channel focus.}
    \label{fig:attention_maps}
\end{center}
\vskip -0.3in
\end{figure*}

\noindent \textbf{Evaluation. }We assess our models, trained on IDRCell100k, to gauge their capacity to generate versatile representations ideal for a range of biologically relevant tasks based on known benchmarking datasets unrelated to the training set, using linear probing, embedding direct evaluation (KNN), as well as image generation for evaluation with 5 different random seeds per run. Classification tasks involved differentiating cell types, protein localizations, and cell cycle stages within the BloodMNIST\cite{Yang2023}, CyclOPS, and BBBC048 datasets, respectively, each of which have varying sizes and complexities. For regression, the NF-kB Nuclear Translocation Assay dataset\cite{lamiable2023revealing} tested the models' ability to quantify protein displacement between cancer cells compartments. The generative task with the BBBC021 dataset\cite{bbbc021} challenged the models to reconstruct cell imaging channels from the encoded \textit{CLS} representations. We froze the encoder and trained a simple convolutional decoder to predict the Actin channel based on other channels for this task. Table \ref{tab:datasets} presents a comprehensive view of the evaluation tasks and dataset details. The One Channel model's evaluation involved using concatenated CLS token representations from each channel for a comprehensive representation. Performance metrics, accuracy for classification and R2 for regression and generation, in addition to Mean Squared Error (MSE) and Mean Absolute Error (MAE) for generation, were selected to measure the models' efficacy precisely. Detailed dataset information and biological context are provided in the appendix.

\section{Results}
\label{sec:results}

\textbf{Biologically Relevant tasks. } Our experimental outcomes, as detailed in Figure \ref{fig:main_results}, indicate that the proposed ChAda-ViT model, with its dual focus on Inter-Channel and Intra-Channel Attention, surpasses the standard one-channel approach in 6 out of the 8 tasks evaluated. This is notable considering the one-channel approach employs a larger representation space when using the CLS token only, as shown in Table \ref{tab:representation_dimension}. Evaluation on all output tokens in Appendix showcases a similar pattern. These results underline the effectiveness of introducing an inter-channel attention mechanism while training on microscopy images and suggest it leads to a more subtle and efficient biological image representation. Additional comparisons on Standard ViT trained on the 3 channel subset of IDRCell100K in the Appendix further validates the added value of training with Inter-channel attention.

\begin{table}[tb]
\centering
\resizebox{\columnwidth}{!}{%
\begin{tabular}{|l|r|r|}
\hline
Dataset & \multicolumn{1}{l|}{One Channel ViT/16 Tiny} & \multicolumn{1}{l|}{ChAda ViT/16 Moy} \\ \hline
BloodMNIST             & 576 & 192 \\ \hline
CyclOPS                & 384 & 192 \\ \hline
BBBC048                & 576 & 192 \\ \hline
NF-kB Nuclear Transloc & 576 & 192 \\ \hline
BBBC021                & 384 & 192 \\ \hline
\end{tabular}%
}
\caption{Comparison of representation dimensions across downstream task datasets for One Channel ViT/16 Tiny and ChAda ViT/16 Moyen using CLS token only. The One Channel ViT, influenced by its dependence on the input channel counts, offers larger and more varied embedding dimensions, theorically leading to more extensive but less channel-interrelated image representations.}
\label{tab:representation_dimension}
\vskip -0.15in
\end{table}

Moreover, the experiments suggest that solely employing Inter-Channel attention, as shown by the Inter-Channel only ViT, might be insufficient for capturing the complexities of biological imaging. Instead, the amalgamation of both Inter-Channel and Intra-Channel Attention, as implemented in ChAda-ViT, yields superior representation quality compared to the application of each method in isolation. This integrative approach harnesses the strengths of both attention mechanisms to enhance the model's performance.

However, in the classification and clustering tasks within the BBBC048 dataset, the One Channel ViT approach, focusing primarily on Intra-Channel Attention, demonstrates a better performance over ChAda-ViT. This outcome could be attributed to the characteristics of the BBBC048 dataset (illustrated in Appendix). The images in this dataset reveal that certain features, crucial for classifying cell cycle stages, are already predominantly present within each single channel. With the representation power of One Channel ViT being bigger than ChAda-ViT Moyen (see Table \ref{tab:representation_dimension}), for the same input size, it performs better at this classification task, diminishing the need for Inter-Channel relationship analysis for accurate classification.

\noindent \textbf{Inter-Channel Attention. }The comparative analysis of the last-layer self-attention maps between One Channel ViT and ChAda-ViT on the IDRCell100k image channels, as depicted in Figure \ref{fig:attention_maps}, reveals significant insights into the models' focus and interpretability. ChAda-ViT's use of Inter-Channel Attention is a pivotal aspect that distinguishes its performance from the One Channel ViT. Specifically, ChAda-ViT demonstrates a heightened ability to establish meaningful correlations across different channels, as indicated by the red arrow. 
This capability allows it to associate biological information from various channels, effectively enhancing focus on spatial locations in a channel that might otherwise appear information-scarce.
Similarly, yellow arrows highlight the ChAda-ViT's strategic focus on spatially relevant areas, such as intra-nuclei features in the yellow-framed images, even in channels with limited information—contrasting with the One Channel ViT's approach.
This targeted approach allows ChAda-ViT to unveil key biological features, which could potentially be missed in a single-channel analysis. Such capability is especially valuable in complex biological imaging where multiple channels convey different but interconnected information.

\begin{figure}[tb!]
\begin{center}
\centerline{\includegraphics[width=\columnwidth]{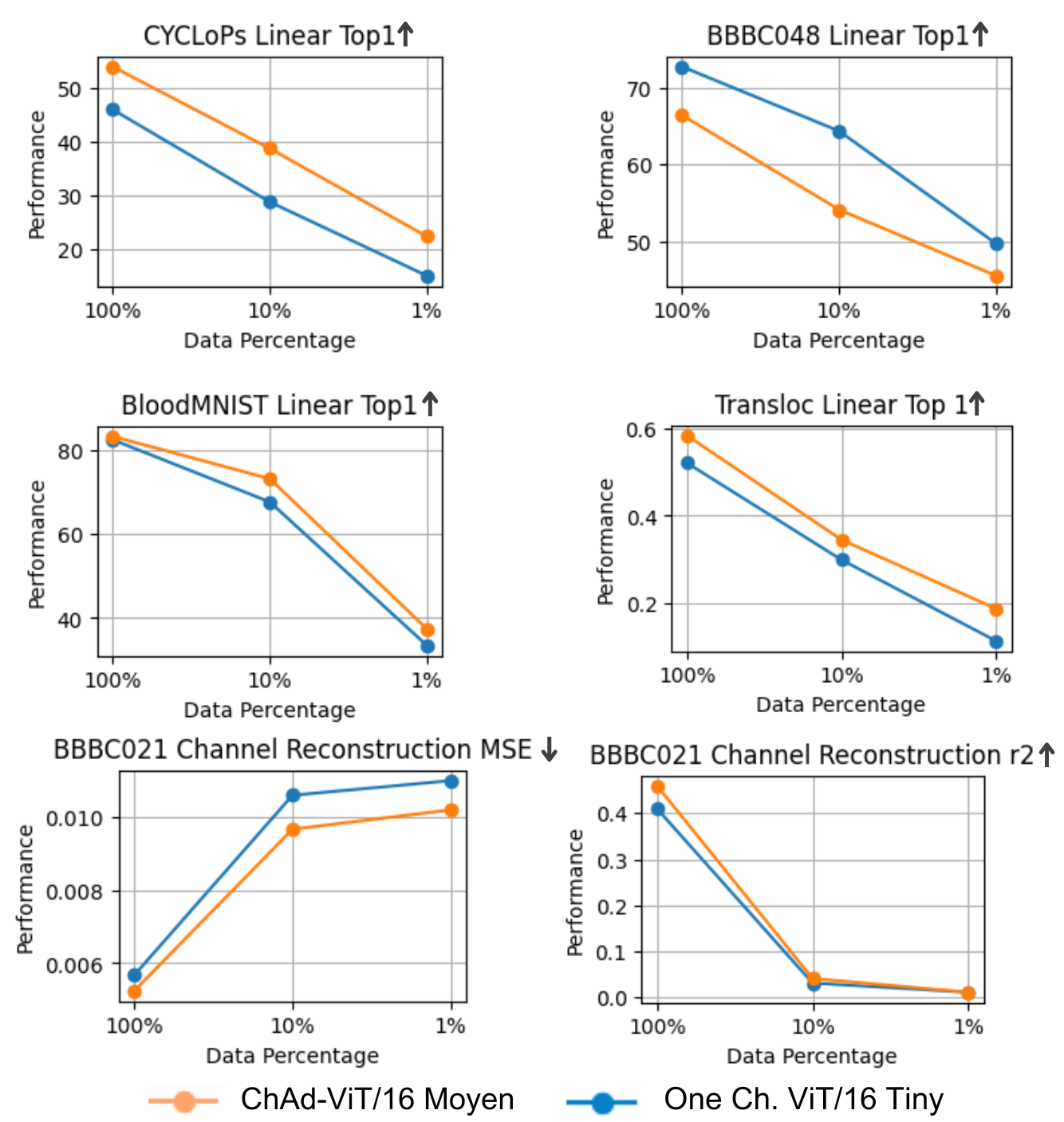}}
    \caption{Evaluation of ChAda-ViT's performance in linear probing low-data regimes, utilizing 100\%, 10\%, and 1\% of the training data for linear probing. The results consistently demonstrate ChAda-ViT's relative performance across tasks, maintaining the performance trends in the same tasks, regardless of the data volume.}
    \label{fig:low_shot_fig}
\end{center}
\vskip -0.3in
\end{figure}

While qualitative, the implications of these findings are substantial in the context of biological image analysis. Evidences in ChAda-ViT's proficiency to recognize and emphasize spatially relevant areas across channels underscores its utility in deciphering complex biological structures and functions. This multi-channel attention helps the model to construct a more holistic and nuanced understanding of the cellular components and their interactions. This intricate understanding could lead to more accurate and comprehensive interpretations of biological data, particularly in scenarios where multiple channels contribute to the overall picture. The attention maps are thus consistent with the effectiveness of ChAda-ViT's design but also offer a window into the model's operational dynamics, highlighting its potential to significantly enhance the analysis of multi-channel biological imaging.

\noindent \textbf{Low Data Regime. }
We then delved into how the performance of ChAda-ViT models compares to one-channel ViT models under constrained data conditions. Specifically, we conducted linear probing using 100\%, 10\%, and 1\% of the available training data for each downstream task. This aspect of the study is crucial for evaluating model behavior in real world laboratory context, where data related to a specific experiment is often limited. Such situations commonly involve either fine-tuning a pre-trained model with the available data or employing the model directly as is.

As illustrated in Figure \ref{fig:low_shot_fig}, our findings reveal that the ChAda-ViT model maintains a consistent performance trend across various data regimes. This persistence in performance, regardless of the amount of data used, is particularly significant. It indicates that once pre-trained, our ChAda-ViT model is robust and capable of achieving high-quality results, even in low-data regime scenarios commonly encountered in biological research. This consistency underscores the practical applicability and reliability of our models in diverse real-world biological research settings.

\begin{figure*}[tb!]
\begin{center}
\centerline{\includegraphics[width=\textwidth]{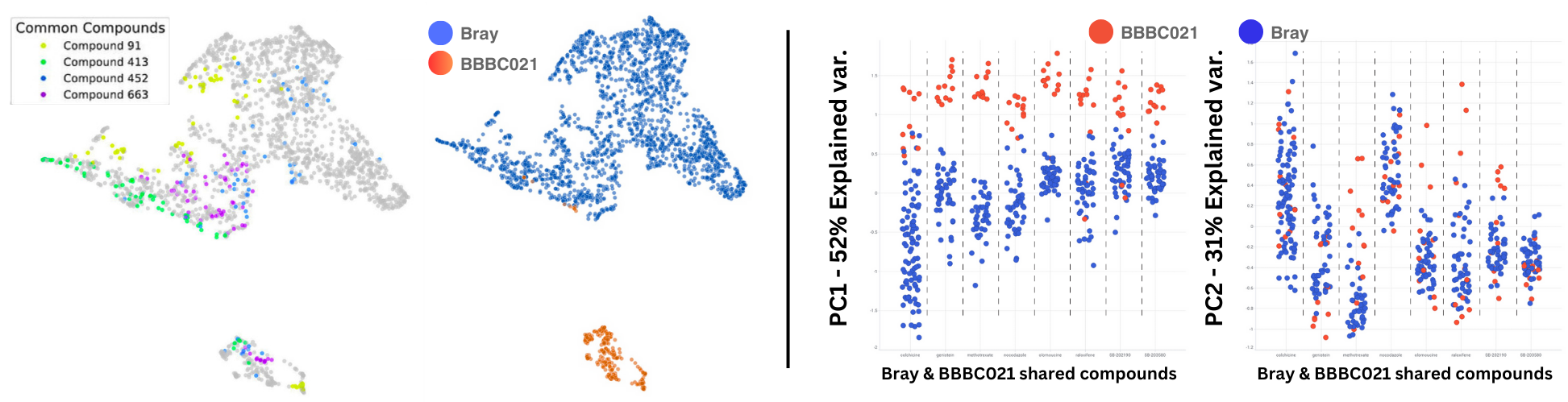}}
    \caption{UMAP \& PCA projections of the BBBC021 (3 channels) and Bray (5 channels) datasets in a unified representation space, derived from ChAda-ViT Moyen. The left UMAP projection highlights the possibility of projecting structurally different experiments into the same space. The right PCA projection and components are labeled by dataset, with a linear separation of the structural differences of different experiments (PC1) and different compounds (PC2) using only the two datasets' common compounds.}
    \label{fig:brayxbbbc}
\end{center}
\vskip -0.3in
\end{figure*}

\noindent \textbf{Single Joint Embedding Space. }In addition, a pivotal contribution of the ChAda-ViT model lies in its ability to unify different datasets, each with distinct channel types and counts, into a consistent embedding space. This feature could end up being particularly advantageous for cross-experimental studies in biology, where diverse experiments often produce sparse data with varying imaging techniques and channels. Such a unified approach would highly benefit fields such as drug discovery, where integrating varied experimental results can provide deeper insights.

For instance, the BBBC021 dataset, composed of 3 channels per image, showcases whole-slide images of cells responding to specific chemical compounds, captured using the Cell Painting technique. Similarly, the Bray dataset\cite{bray_data}, composed of 5 channels per image, as well as different experimental conditions, displays cells affected by various compounds, but imaged with a higher channel counts encoding distinct biological information than the first dataset. Notably, these datasets share at least 22 common compound treatments that could benefit from cross-dataset comparisons. By examining these shared compound treatments, it might be possible shed light on the interconnection of different mechanisms of action and understand the broader impact of these chemical compounds. However, existing approaches are unable to properly compare these different experimental settings, due different learnt embedding spaces caused by differences in channel count. This common representation space, uniquely facilitated by ChAda-ViT architecture, allows us to leverage data from other compounds or experiments, providing a bridge between numerous but disconnected biological datasets currently available. This approach could potentially enhance our understanding of a wider range of compounds, serving as a preliminary step that capitalizes on the variety of open experiments in this domain, and laying the groundwork for more extensive future research.

Figure \ref{fig:brayxbbbc} presents a UMAP projection of these two datasets, highlighting their common compounds, within the same representation (left), achieved through ChAda-ViT Moyen, while similarly highlighting that the model, trained on only 100k images before projecting the representatons of the two datasets' common compound using a PCA, can linearly separate the two datasets through only the PCA first Principal Component (PC1), while differentiating the different compounds from each other in the PCA second component (PC2) (right). While aligning various labels (such as compound and cell types) in the representation is necessary for some cross-dataset tasks, this shared representation space is a significant initial step. It serves as a foundation for further exploration and potential breakthroughs in cross-experimental biological research.

\section{Conclusion}
\label{sec:conclusion}

In this study, we introduced ChAda-ViT, a Channel-Adaptive Vision Transformer designed specifically for multichannel image data. ChAda-ViT integrates token padding, masking as well as channel and positional embeddings within a patch-based framework, making it highly effective for handling diverse types of imaging data. On top of the added inter-channel attention, the standout feature of ChAda-ViT is the ability to bring disparate microscopy images into a single, joint embedding space, facilitating comprehensive comparisons across varied datasets with distinct types, channel counts, and imaging techniques. 

Our model demonstrates superior performance over existing methods in the literature across most tasks, including normal and low data regime linear probing. It also provides more insightful attention maps at the channel level, paving the way for new explorations into cellular components and their interactions. The potential of ChAda-ViT extends to aligning various biological datasets within its embedding space, offering novel avenues for cross-experimental studies and new biological insights. This capability offers a significant potential to the field of biological image analysis, e.g. leading to augment datasets with additional channels based on learnings from other datasets, thereby reducing experimental costs.

Another key contribution of our work is the introduction of the IDRCell100k dataset, a first-of-its-kind collection featuring heterogeneous biological image sources and a range of multi-channel configurations. This dataset not only demonstrates the adaptability and robustness of ChAda-ViT but also serves as a critical asset for ongoing and future research in representation learning for biological images.

Nevertheless, our research is not without limitations. Although ChAda-ViT excels in many tasks, it falls short in tasks heavily reliant on intra-channel information. Addressing this shortfall could position our approach as the go-to method for encoding biological images, encapsulating both inter- and intra-channel dynamics. Furthermore, exploring ChAda-ViT as basis for a foundation model, potentially leveraging massive datasets, could unveil new capabilities and applications. Bridging this gap and expanding its dataset foundation remain key objectives for future enhancements, with the aim of solidifying ChAda-ViT's role as a cornerstone technology in biological imaging analysis.

\section{Acknowledgments}
\label{sec:acknowledgments}
This work was supported by ANR–10–LABX–54 MEMOLIFE and ANR–10 IDEX 0001-02 PSL* Université Paris and was granted access to the HPC resources of IDRIS under the allocations 2020-AD011011495 and 2023-AD011014487 made by GENCI.
{
    \small
    \bibliographystyle{ieeenat_fullname}
    \bibliography{main}

\begin{thebibliography}{51}
\providecommand{\natexlab}[1]{#1}
\providecommand{\url}[1]{\texttt{#1}}
\expandafter\ifx\csname urlstyle\endcsname\relax
  \providecommand{\doi}[1]{doi: #1}\else
  \providecommand{\doi}{doi: \begingroup \urlstyle{rm}\Url}\fi

\bibitem[Allier et~al.(2021)Allier, Hervé, Paviolo, Mandula, Cioni, Pierre, Andriani, Padmanabhan, and Morales]{frontierscnn}
Cédric Allier, Lionel Hervé, Chiara Paviolo, Ondrej Mandula, Olivier Cioni, William Pierre, Francesca Andriani, Kiran Padmanabhan, and Sophie Morales.
\newblock Cnn-based cell analysis: From image to quantitative representation.
\newblock \emph{Frontiers in Bioengineering and Biotechnology}, 9:\penalty0 673840, 2021.

\bibitem[Assran et~al.(2023)Assran, Duval, Misra, Bojanowski, Vincent, Rabbat, LeCun, and Ballas]{assran2023selfsupervised}
Mahmoud Assran, Quentin Duval, Ishan Misra, Piotr Bojanowski, Pascal Vincent, Michael Rabbat, Yann LeCun, and Nicolas Ballas.
\newblock Self-supervised learning from images with a joint-embedding predictive architecture.
\newblock In \emph{ICCV}, 2023.

\bibitem[Bendidi et~al.(2023)Bendidi, Bardes, Cohen, Lamiable, Bollot, and Genovesio]{nofreelunch}
Ihab Bendidi, Adrien Bardes, Ethan Cohen, Alexis Lamiable, Guillaume Bollot, and Auguste Genovesio.
\newblock No free lunch in self supervised representation learning, 2023.

\bibitem[Bourou and et~al.(2023)]{bourou2023unpaired}
Anis Bourou and K{\'e}vin~Daupin et al.
\newblock Unpaired image-to-image translation with limited data to reveal subtle phenotypes.
\newblock In \emph{2023 IEEE 20th International Symposium on Biomedical Imaging (ISBI)}, pages 1--5. IEEE, 2023.

\bibitem[Bray and Gustafsdottir(2017)]{bray_data}
Mark-Anthony Bray and Sigrun M et~al. Gustafsdottir.
\newblock {A dataset of images and morphological profiles of 30 000 small-molecule treatments using the Cell Painting assay}.
\newblock \emph{GigaScience}, 6\penalty0 (12):\penalty0 giw014, 2017.

\bibitem[Caie et~al.(2010)Caie, Walls, and et~al.]{bbbc021}
Peter~D. Caie, Rebecca~E. Walls, and Alexandra Ingleston-Orme et al.
\newblock {High-Content Phenotypic Profiling of Drug Response Signatures across Distinct Cancer Cells}.
\newblock \emph{Molecular Cancer Therapeutics}, 9\penalty0 (6):\penalty0 1913--1926, 2010.

\bibitem[Carion et~al.(2020)Carion, Massa, Synnaeve, Usunier, Kirillov, and Zagoruyko]{Carion2020EndtoEndOD}
Nicolas Carion, Francisco Massa, Gabriel Synnaeve, Nicolas Usunier, Alexander Kirillov, and Sergey Zagoruyko.
\newblock End-to-end object detection with transformers.
\newblock In \emph{ECCV}, 2020.

\bibitem[Caron et~al.(2021)Caron, Touvron, Misra, Jégou, Mairal, Bojanowski, and Joulin]{dino}
Mathilde Caron, Hugo Touvron, Ishan Misra, Hervé Jégou, Julien Mairal, Piotr Bojanowski, and Armand Joulin.
\newblock Emerging properties in self-supervised vision transformers.
\newblock In \emph{ICCV}, 2021.

\bibitem[Chandrasekaran and et~al.(2023)]{cell_painting}
Srinivas~Niranj Chandrasekaran and Jeanelle~Ackerman et al.
\newblock Jump cell painting dataset: morphological impact of 136,000 chemical and genetic perturbations.
\newblock \emph{bioRxiv}, 2023.

\bibitem[Chandrasekaran et~al.(2022)Chandrasekaran, Cimini, and et~al.]{Chandrasekaran2022ThreeMI}
Srinivas~Niranj Chandrasekaran, Beth~A. Cimini, and Amy~Goodale et al.
\newblock Three million images and morphological profiles of cells treated with matched chemical and genetic perturbations.
\newblock \emph{bioRxiv}, 2022.

\bibitem[Chen et~al.(2021)Chen, Lu, Yu, Luo, Adeli, Wang, Lu, Yuille, and Zhou]{Chen2021TransUNetTM}
Jieneng Chen, Yongyi Lu, Qihang Yu, Xiangde Luo, Ehsan Adeli, Yan Wang, Le Lu, Alan~Loddon Yuille, and Yuyin Zhou.
\newblock Transunet: Transformers make strong encoders for medical image segmentation.
\newblock \emph{ArXiv}, abs/2102.04306, 2021.

\bibitem[Cohen and Uhlmann(2021)]{Cohen2021AuraNetRS}
Ethan Cohen and Virginie Uhlmann.
\newblock Aura-net: Robust segmentation of phase-contrast microscopy images with few annotations.
\newblock \emph{2021 IEEE 18th International Symposium on Biomedical Imaging (ISBI)}, pages 640--644, 2021.

\bibitem[Cohen et~al.(2022)Cohen, Corb{\'e}, Franco, Vasconcelos, Perez, Nery, Bollot, and Genovesio]{Cohen2022CellPT}
Ethan Cohen, Maxime Corb{\'e}, Cl{\'a}udio~Areias Franco, Francisca~F. Vasconcelos, Franck Perez, Elaine~Del Nery, Guillaume Bollot, and Auguste Genovesio.
\newblock Cell painting transfer increases screening hit rate.
\newblock \emph{bioRxiv}, 2022.

\bibitem[Cross-Zamirski et~al.(2021)Cross-Zamirski, Mouchet, Williams, Sch{\"o}nlieb, Turkki, and Wang]{CrossZamirski2021LabelfreePO}
Jan~Oscar Cross-Zamirski, Elizabeth Mouchet, Guy~B. Williams, Carola-Bibiane Sch{\"o}nlieb, Riku Turkki, and Yinhai Wang.
\newblock Label-free prediction of cell painting from brightfield images.
\newblock \emph{Scientific Reports}, 12, 2021.

\bibitem[Dai et~al.(2020)Dai, Cai, Lin, and Chen]{Dai2020UPDETRUP}
Zhigang Dai, Bolun Cai, Yugeng Lin, and Junying Chen.
\newblock Up-detr: Unsupervised pre-training for object detection with transformers.
\newblock \emph{2021 IEEE/CVF Conference on Computer Vision and Pattern Recognition (CVPR)}, pages 1601--1610, 2020.

\bibitem[Dosovitskiy et~al.(2020)Dosovitskiy, Beyer, Kolesnikov, Weissenborn, Zhai, Unterthiner, Dehghani, Minderer, Heigold, Gelly, Uszkoreit, and Houlsby]{Dosovitskiy2020AnII}
Alexey Dosovitskiy, Lucas Beyer, Alexander Kolesnikov, Dirk Weissenborn, Xiaohua Zhai, Thomas Unterthiner, Mostafa Dehghani, Matthias Minderer, Georg Heigold, Sylvain Gelly, Jakob Uszkoreit, and Neil Houlsby.
\newblock An image is worth 16x16 words: Transformers for image recognition at scale.
\newblock \emph{ArXiv}, abs/2010.11929, 2020.

\bibitem[et~al.(2023{\natexlab{a}})]{Doron2023UnbiasedSM}
Michael~Doron et al.
\newblock Unbiased single-cell morphology with self-supervised vision transformers.
\newblock \emph{bioRxiv}, 2023{\natexlab{a}}.

\bibitem[et~al.(2023{\natexlab{b}})]{oquab2023dinov2}
Maxime~Oquab et al.
\newblock Dinov2: Learning robust visual features without supervision, 2023{\natexlab{b}}.

\bibitem[Filiot et~al.(2023)Filiot, Ghermi, Olivier, Jacob, Fidon, Kain, Saillard, and Schiratti]{Filiot2023scalingwithMIM}
Alexandre Filiot, Ridouane Ghermi, Antoine Olivier, Paul Jacob, Lucas Fidon, Alice~Mac Kain, Charlie Saillard, and Jean-Baptiste Schiratti.
\newblock Scaling self-supervised learning for histopathology with masked image modeling.
\newblock \emph{medRxiv}, 2023.

\bibitem[Gabriel et~al.(2023)Gabriel, Ethan, Nicolas, Ihab, Guillaume, and Auguste]{gabriel2023weakly}
Watkinson Gabriel, Cohen Ethan, Bourriez Nicolas, Bendidi Ihab, Bollot Guillaume, and Genovesio Auguste.
\newblock Weakly supervised cross-model learning in high-content screening, 2023.

\bibitem[Goltsev et~al.(2017)Goltsev, Samusik, Kennedy‐Darling, Bhate, Hale, Vazquez, Black, and Nolan]{Goltsev2017DeepPO}
Yury Goltsev, Nikolay Samusik, Julia Kennedy‐Darling, Salil~S. Bhate, Matthew~B. Hale, Gustavo Vazquez, Sarah Black, and Garry~P. Nolan.
\newblock Deep profiling of mouse splenic architecture with codex multiplexed imaging.
\newblock \emph{Cell}, 174:\penalty0 968 -- 981.e15, 2017.

\bibitem[Harkness et~al.(2022)Harkness, Theis, O'Keefe, Wade, and Lawton]{Harkness2022LeveragingAT}
John~H Harkness, Jacob~J. Theis, Will~M. O'Keefe, Grant~W. Wade, and Kristy~J. Lawton.
\newblock Leveraging ai transfer learning for rapid and accurate identification and quantification of cellular biomarkers in microscopy images.
\newblock \emph{The FASEB Journal}, 36, 2022.

\bibitem[Hartley et~al.(2021)Hartley, Kleywegt, Patwardhan, Sarkans, Swedlow, and Brazma]{Hartley2021TheBA}
Matthew Hartley, Gerard~J. Kleywegt, Ardan Patwardhan, Ugis Sarkans, Jason~R. Swedlow, and Alvis Brazma.
\newblock The bioimage archive - building a home for life-sciences microscopy data.
\newblock \emph{bioRxiv}, 2021.

\bibitem[He et~al.(2015)He, Zhang, Ren, and Sun]{He2015DeepRL}
Kaiming He, X. Zhang, Shaoqing Ren, and Jian Sun.
\newblock Deep residual learning for image recognition.
\newblock \emph{2016 IEEE Conference on Computer Vision and Pattern Recognition (CVPR)}, pages 770--778, 2015.

\bibitem[He et~al.(2021)He, Chen, Xie, Li, Dollár, and Girshick]{he2021masked}
Kaiming He, Xinlei Chen, Saining Xie, Yanghao Li, Piotr Dollár, and Ross Girshick.
\newblock Masked autoencoders are scalable vision learners, 2021.

\bibitem[Hervé et~al.(2020)Hervé, Kraemer, Cioni, Menneteau, Morales, and Allier]{cnn_lensfree}
L. Hervé, D.~C.~A. Kraemer, O. Cioni, O.and~Mandula, M. Menneteau, S. Morales, and C. Allier.
\newblock Alternation of inverse problem approach and deep learning for lens-free microscopy image reconstruction.
\newblock \emph{Scientific Reports}, 10\penalty0 (1):\penalty0 20207, 2020.

\bibitem[Hickey et~al.(2021)Hickey, Neumann, and et~al.]{Hickey2021SpatialMO}
John~W. Hickey, Elizabeth~K. Neumann, and Andrea J.~Radtke et al.
\newblock Spatial mapping of protein composition and tissue organization: a primer for multiplexed antibody-based imaging.
\newblock \emph{Nature Methods}, 19:\penalty0 284 -- 295, 2021.

\bibitem[Hua et~al.(2021)]{cytoimagenet}
Stanley Hua et~al.
\newblock Cytoimagenet: A large-scale pretraining dataset for bioimage transfer learning.
\newblock In \emph{NeuIPS LMRL Workshop}, 2021.

\bibitem[Kim et~al.(2023)Kim, Adaloglou, Osterland, Morelli, and Zapata]{kim2023selfsupervision}
Vladislav Kim, Nikolaos Adaloglou, Marc Osterland, Flavio~M Morelli, and Paula A~Marin Zapata.
\newblock Self-supervision advances morphological profiling by unlocking powerful image representations.
\newblock \emph{bioRxiv}, 2023.

\bibitem[Krizhevsky et~al.(2012)Krizhevsky, Sutskever, and Hinton]{Krizhevsky2012ImageNetCW}
Alex Krizhevsky, Ilya Sutskever, and Geoffrey~E. Hinton.
\newblock Imagenet classification with deep convolutional neural networks.
\newblock \emph{Communications of the ACM}, 60:\penalty0 84 -- 90, 2012.

\bibitem[Lamiable et~al.(2023)Lamiable, Champetier, Leonardi, Cohen, Sommer, Hardy, Argy, Massougbodji, Del~Nery, Cottrell, et~al.]{lamiable2023revealing}
Alexis Lamiable, Tiphaine Champetier, Francesco Leonardi, Ethan Cohen, Peter Sommer, David Hardy, Nicolas Argy, Achille Massougbodji, Elaine Del~Nery, Gilles Cottrell, et~al.
\newblock Revealing invisible cell phenotypes with conditional generative modeling.
\newblock \emph{Nature Communications}, 14\penalty0 (1):\penalty0 6386, 2023.

\bibitem[LeCun and et~al.(1989)]{LeCun1989BackpropagationAT}
Yann LeCun and Bernhard E.~Boser et al.
\newblock Backpropagation applied to handwritten zip code recognition.
\newblock \emph{Neural Computation}, 1:\penalty0 541--551, 1989.

\bibitem[Li et~al.(2023)Li, Zhang, Wu, and Dai]{li2023challenges}
Xinyang Li, Yuanlong Zhang, Jiamin Wu, and Qionghai Dai.
\newblock Challenges and opportunities in bioimage analysis.
\newblock \emph{Nature Methods}, 20\penalty0 (4):\penalty0 367--377, 2023.
\newblock \url{https://www.nature.com/articles/s41592-023-01900-4}.

\bibitem[Masud et~al.(2022)Masud, Cohen, Bendidi, Bollot, and Genovesio]{Masud2022ComparisonOS}
Umar Masud, Ethan~O. Cohen, Ihab Bendidi, Guillaume Bollot, and Auguste Genovesio.
\newblock Comparison of semi-supervised learning methods for high content screening quality control.
\newblock In \emph{ECCV Workshops}, 2022.

\bibitem[McQuin et~al.(2018)McQuin, Goodman, Chernyshev, Kamentsky, Cimini, Karhohs, et~al.]{mcquin2018cellprofiler}
Claire McQuin, Allen Goodman, Vasiliy Chernyshev, Lee Kamentsky, Beth~A Cimini, Kyle~W Karhohs, et~al.
\newblock Cellprofiler 3.0: Next-generation image processing for biology.
\newblock \emph{PLoS Biol}, 16\penalty0 (7):\penalty0 e2005970, 2018.

\bibitem[Meng et~al.(2021)Meng, Li, Chen, Lan, Wu, Jiang, and Lim]{Meng2021AdaViTAV}
Lingchen Meng, Hengduo Li, Bor-Chun Chen, Shiyi Lan, Zuxuan Wu, Yu-Gang Jiang, and Ser~Nam Lim.
\newblock Adavit: Adaptive vision transformers for efficient image recognition.
\newblock \emph{2022 IEEE/CVF Conference on Computer Vision and Pattern Recognition (CVPR)}, pages 12299--12308, 2021.

\bibitem[Moshkov et~al.(2023)Moshkov, Becker, Yang, Horvath, Dancik, Wagner, Clemons, Singh, Carpenter, and Caicedo]{Moshkov2023-jg}
Nikita Moshkov, Tim Becker, Kevin Yang, Peter Horvath, Vlado Dancik, Bridget~K Wagner, Paul~A Clemons, Shantanu Singh, Anne~E Carpenter, and Juan~C Caicedo.
\newblock Predicting compound activity from phenotypic profiles and chemical structures.
\newblock \emph{Nature Communications}, 14\penalty0 (1):\penalty0 1967, 2023.

\bibitem[Rao et~al.(2021)Rao, Zhao, Liu, Lu, Zhou, and Hsieh]{Rao2021DynamicViTEV}
Yongming Rao, Wenliang Zhao, Benlin Liu, Jiwen Lu, Jie Zhou, and Cho-Jui Hsieh.
\newblock Dynamicvit: Efficient vision transformers with dynamic token sparsification.
\newblock \emph{ArXiv}, abs/2106.02034, 2021.

\bibitem[Sch{\"o}lkopf et~al.(2021)Sch{\"o}lkopf, Locatello, Bauer, Ke, Kalchbrenner, Goyal, and Bengio]{Schlkopf2021TowardCR}
Bernhard Sch{\"o}lkopf, Francesco Locatello, Stefan Bauer, Nan~Rosemary Ke, Nal Kalchbrenner, Anirudh Goyal, and Yoshua Bengio.
\newblock Toward causal representation learning.
\newblock \emph{Proceedings of the IEEE}, 109:\penalty0 612--634, 2021.

\bibitem[Sha and Li(2022)]{Sha2022MITformerAM}
Zongyao Sha and Jianfeng Li.
\newblock Mitformer: A multiinstance vision transformer for remote sensing scene classification.
\newblock \emph{IEEE Geoscience and Remote Sensing Letters}, 19:\penalty0 1--5, 2022.

\bibitem[Thain et~al.(2005)Thain, Tannenbaum, and Livny]{condor-practice}
Douglas Thain, Todd Tannenbaum, and Miron Livny.
\newblock Distributed computing in practice: the condor experience.
\newblock \emph{Concurrency - Practice and Experience}, 17\penalty0 (2-4):\penalty0 323--356, 2005.

\bibitem[Touvron et~al.(2021)Touvron, Cord, Douze, Massa, Sablayrolles, and J{\'e}gou]{touvron2020training}
Hugo Touvron, Matthieu Cord, Matthijs Douze, Francisco Massa, Alexandre Sablayrolles, and Herv{\'e} J{\'e}gou.
\newblock Training data-efficient image transformers \& distillation through attention.
\newblock \emph{PMLR}, 2021.

\bibitem[Vaswani et~al.(2017)Vaswani, Shazeer, Parmar, Uszkoreit, Jones, Gomez, Kaiser, and Polosukhin]{Vaswani2017AttentionIA}
Ashish Vaswani, Noam~M. Shazeer, Niki Parmar, Jakob Uszkoreit, Llion Jones, Aidan~N. Gomez, Lukasz Kaiser, and Illia Polosukhin.
\newblock Attention is all you need.
\newblock In \emph{Neural Information Processing Systems}, 2017.

\bibitem[von Chamier et~al.(2021)von Chamier, Laine, and et~al.]{vonChamier2021DemocratisingDL}
Lucas von Chamier, Romain~F. Laine, and Johanna~Jukkala et al.
\newblock Democratising deep learning for microscopy with zerocostdl4mic.
\newblock \emph{Nature Communications}, 12, 2021.

\bibitem[Williams and Moore(2017)]{Williams2017}
Eleanor Williams and Josh et~al. Moore.
\newblock Image data resource: a bioimage data integration and publication platform.
\newblock \emph{Nature Methods}, 14\penalty0 (8):\penalty0 775--781, 2017.

\bibitem[Xie et~al.(2021)Xie, Wang, Yu, Anandkumar, {\'A}lvarez, and Luo]{Xie2021SegFormerSA}
Enze Xie, Wenhai Wang, Zhiding Yu, Anima Anandkumar, Jos{\'e}~Manuel {\'A}lvarez, and Ping Luo.
\newblock Segformer: Simple and efficient design for semantic segmentation with transformers.
\newblock \emph{ArXiv}, abs/2105.15203, 2021.

\bibitem[Xun et~al.(2023)Xun, Wang, and Wang]{microsnoop}
Dejin Xun, Rui Wang, and Yi Wang.
\newblock Microsnoop: a generalist tool for the unbiased representation of heterogeneous microscopy images.
\newblock \emph{bioRxiv}, 2023.

\bibitem[Yang et~al.(2023)Yang, Shi, Wei, Liu, Zhao, Ke, Pfister, and Ni]{Yang2023}
Jiancheng Yang, Rui Shi, Donglai Wei, Zequan Liu, Lin Zhao, Bilian Ke, Hanspeter Pfister, and Bingbing Ni.
\newblock Medmnist v2 - a large-scale lightweight benchmark for 2d and 3d biomedical image classification.
\newblock \emph{Scientific Data}, 10\penalty0 (1):\penalty0 41, 2023.

\bibitem[Yu et~al.(2021)Yu, Ma, Bi, Bian, Ning, He, Li, Liu, and Zheng]{Yu2021MILVTMI}
Shuang Yu, Kai Ma, Qi Bi, Cheng Bian, Munan Ning, Nanjun He, Yuexiang Li, Hanruo Liu, and Yefeng Zheng.
\newblock Mil-vt: Multiple instance learning enhanced vision transformer for fundus image classification.
\newblock In \emph{International Conference on Medical Image Computing and Computer-Assisted Intervention}, 2021.

\bibitem[Yuan et~al.(2021)Yuan, Chen, Wang, Yu, Shi, Tay, Feng, and Yan]{Yuan2021TokenstoTokenVT}
Li Yuan, Yunpeng Chen, Tao Wang, Weihao Yu, Yujun Shi, Francis E.~H. Tay, Jiashi Feng, and Shuicheng Yan.
\newblock Tokens-to-token vit: Training vision transformers from scratch on imagenet.
\newblock \emph{2021 IEEE/CVF International Conference on Computer Vision (ICCV)}, pages 538--547, 2021.

\bibitem[Zheng et~al.(2022)Zheng, Rao, Zhang, Cohen, Li, and Yang]{zheng2022cross}
Shuangjia Zheng, Jiahua Rao, Jixian Zhang, Ethan Cohen, Chengtao Li, and Yuedong Yang.
\newblock Cross-modal graph contrastive learning with cellular images.
\newblock \emph{bioRxiv}, pages 2022--06, 2022.

\end{thebibliography}
}


\appendix


\clearpage
\setcounter{page}{1}
\maketitlesupplementary

\section{IDRCell100k Construction}
\label{sup_sec:dataset}

We acquired the IDRCell100K dataset using a distributed High-Performance Computing (HPC) cluster managed by HTCondor. Employing multi-processing and multi-threading techniques, the dataset was efficiently downloaded over two weeks. IDRCell100K serves as an initial, diverse channel configuration dataset, a pioneering resource in this domain to the best of our knowledge. Although currently small in scale, it provides an excellent basis for self-supervised learning applications, including training on high-quality microscopy images and evaluating channel-adaptive architectures with a multichannel dataset.

\begin{figure}[h]
\vskip 0.1in
    \centering
    \includegraphics[scale=0.45]{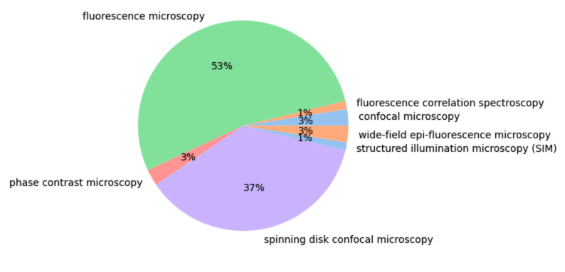}
    \caption{The imaging method distribution within IDRCell-100k, highlighting the dataset's variety in microscopy techniques. This diversity is crucial for developing models capable of interpreting cells imaged under various conditions, despite certain methods being less represented due to higher acquisition costs and technical complexities.}
    \label{fig:idrcell_distribution}
    \vskip -0.1in
\end{figure}

To ensure a broad data distribution, we randomly selected 1,300 images from each study with at least this amount of data available. These images were uniformly chosen from the 8,050,408 images in the "Cell" section of the Image Data Resources datalake. The curated dataset comprises 104,093 multiplexed images, corresponding to 308,898 channel, and features 1 to 10 channels per image. IDRCell100k encompasses microscopy images from 79 different assays and 7 distinct imaging methods (see Fig. \ref{fig:idrcell_distribution}). It will be made freely accessible under an open license, and would require 183GB of disk space for storage.


\section{Evaluation Tasks and Datasets}
\label{sup_sec:evaluation_data}



\subsection*{CYCLoPs}

\textbf{Dataset Description.} The CYCLoPs (Collection of Yeast Cells Localization Patterns) dataset\footnote{The dataset can be accessed on Kaggle: \href{https://www.kaggle.com/datasets/stanleyhua/cyclops-protein-loc}{CYCLoPs Dataset}.} is a specialized collection of 27,058 single-cell yeast images, designed to support research in eukaryotic cell biology. This dataset amalgamates systematic genetics, high-throughput microscopy, and image analysis to reveal protein interactions within cells. A distinctive feature of CYCLoPs is its dual-channel imaging: one channel highlights the protein of interest, and the other visualizes the cytosol. This setup is advantageous for precise visualization and subsequent computational analysis. The dataset's standardization and detailed annotations facilitate its application in machine learning, particularly in deep learning-based classification tasks, without necessitating extensive prior domain knowledge.

\begin{figure}[htp!]
\vskip 0.1in
    \includegraphics[width=\columnwidth]{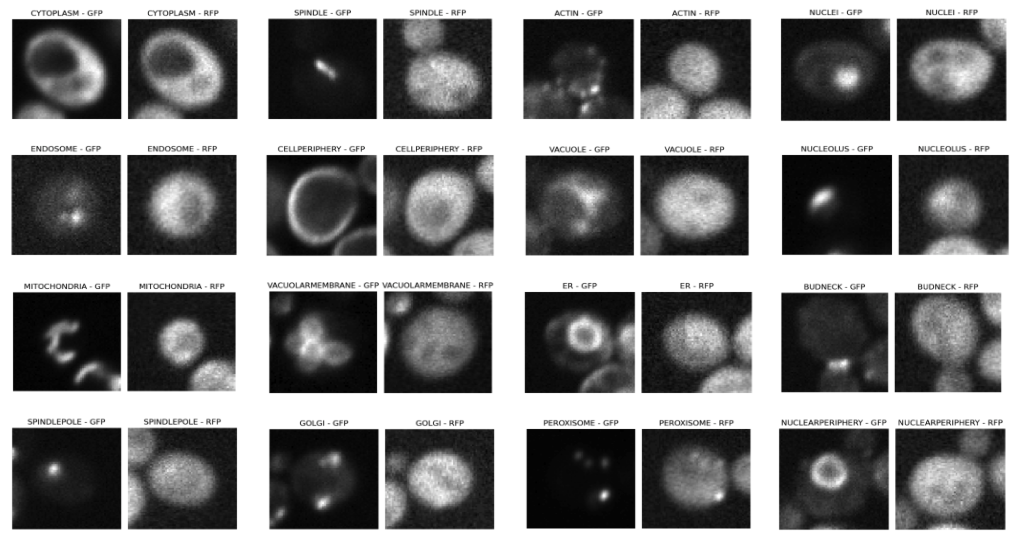}
    \centering
    \caption{The CYCLoPs dataset: A depiction of 16 distinct classes representing diverse protein localizations within the yeast cell, each class corresponding to a unique subcellular region.}
    \label{fig:cyclops}
    \vskip -0.1in
\end{figure}

\textbf{Task Description.} The primary analytical task with the CYCLoPs dataset is the classification of proteins' subcellular localizations in yeast cells. This task is critical for understanding protein network dynamics and cellular functions. Accurate classification of protein localizations provides insights into their roles and interactions within the cell, essential for advancing knowledge in cellular biology and proteomics. The dataset's high-quality, dual-channel images enable precise localization, making it a valuable resource for developing and testing deep learning models for protein localization prediction in eukaryotic cells.

\begin{figure*}[tb]
\vskip 0.1in
\begin{center}
\centerline{\includegraphics[width=\textwidth]{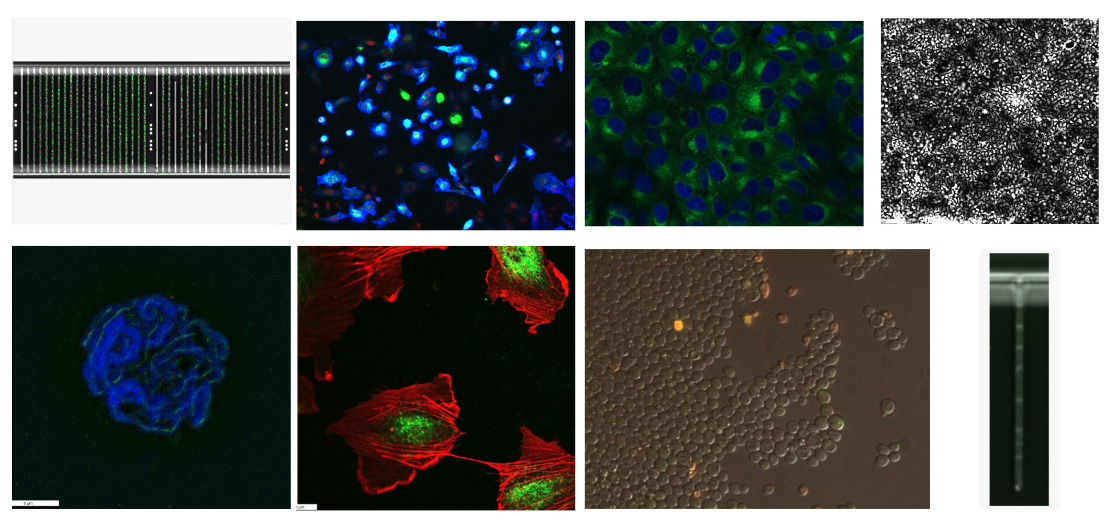}}
    \caption{A selective overview of the diverse microscopy image types in IDRCell-100k, illustrating the dataset's variety. The aim is to achieve a broad and heterogeneous distribution of training data.}
    \label{fig:idrcell100k}
\end{center}
\vskip -0.1in
\end{figure*}

\subsection*{BBBC048}

\textbf{Dataset Description.} The BBBC048 dataset\footnote{More details on the dataset are found here : \href{https://bbbc.broadinstitute.org/BBBC048/}{BBBC048}.}, part of the Broad Bioimage Benchmark Collection, comprises 32,266 images of Jurkat cells, a type of human immune cell. These cells were grown asynchronously and imaged using the ImageStream platform. The dataset's uniqueness lies in its dual staining approach: Propidium Iodide for DNA content quantification and MPM2 antibody for identifying cells in the mitotic phase of the cell cycle. Created by the Flow Cytometry Core Facility at Newcastle University, this dataset is specifically designed to support cell cycle reconstruction and disease progression analysis, particularly through deep learning methods.

\begin{figure}[h!]
\vskip 0.1in
    \includegraphics[width=\columnwidth]{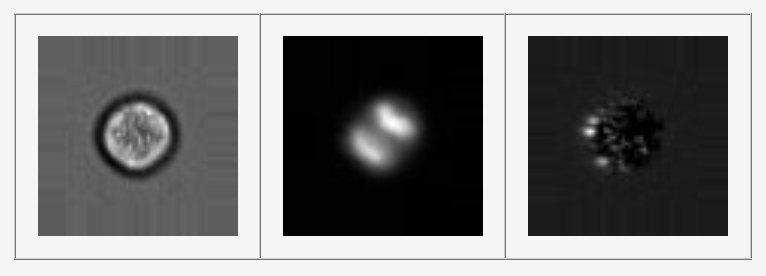}
    \centering
    \caption{Representative Image from BBBC048: A Jurkat cell stained with Propidium Iodide and MPM2 antibody, exemplifying the dataset's dual-staining technique for cell cycle analysis.}
    \label{fig:bbbc048}
    \vskip -0.1in
\end{figure}

\textbf{Task Description.} The primary task with the BBBC048 dataset is to classify discrete stages of the cell cycle. This task is crucial for understanding cellular dynamics and disease progression, particularly in the context of cancer research and other biological processes. The dataset has demonstrated its utility in this regard by achieving a sixfold reduction in error rate for cell cycle stage classification compared to previous boosting-based approaches. Accurate cell cycle stage classification using deep learning models not only advances our understanding of cellular mechanisms but also has potential applications in therapeutic interventions. The BBBC048 dataset thus serves as a pivotal resource for developing robust and generalizable models for cell cycle prediction and analysis from raw image data.

\begin{figure}[ht]
\vskip 0.1in
\begin{center}
\centerline{\includegraphics[width=\columnwidth]{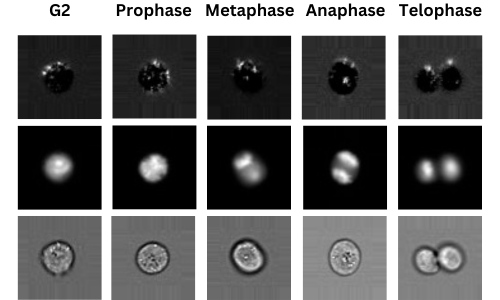}}
    \caption{Randomly sampled BBBC048 images across various Cell Cycle Stages. Each column represents a multiplexed image from a distinct cell cycle, with each row corresponding to a different channel for the same cell. The visualization suggests  specific channels (predominantly the middle row) and their Intra-Channel interactions is sufficient for effective cell cycle classification, lowering down the need to extract Inter-Channel information.}
    \label{fig:cell_cycle_one_channel}
\end{center}
\vskip -0.1in
\end{figure}

\subsection*{BloodMNIST}

\begin{figure}[h!]
\vskip 0.1in
    \includegraphics[width=\columnwidth]{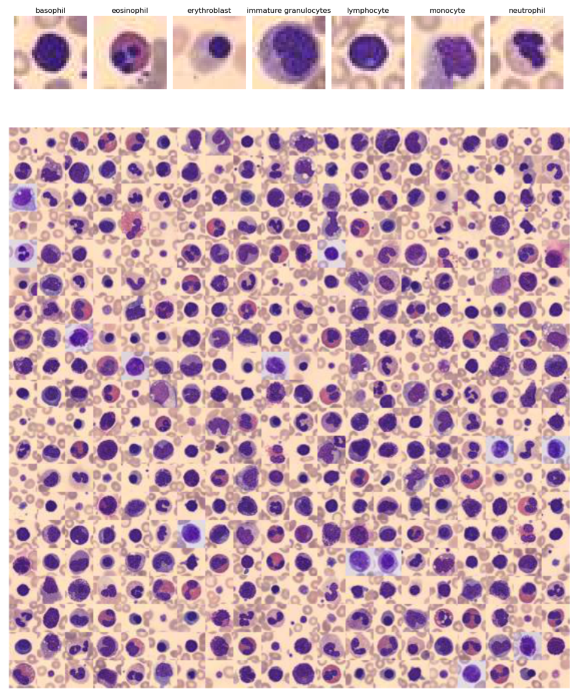}
    \centering
    \caption{Sample Observations from BloodMNIST: The top row displays the 7 different classes of normal blood cells, showcasing the dataset's variety in cellular types.}
    \label{fig:bloodmnist}
    \vskip -0.1in
\end{figure}

\textbf{Dataset Description.} BloodMNIST, a part of the MedMNIST collection\cite{Yang2023}, is a dataset focused on images of normal blood cells. It was created from blood samples collected from individuals free of infections, hematologic or oncologic diseases, and not undergoing any pharmacologic treatments. The dataset contains a total of 17,092 images, which are divided into 7 classes. Each class represents a different type of normal cell found in human blood. The original high-resolution images (3 × 360 × 363 pixels) (see Fig.\ref{fig:bloodmnist}) have been resized to fit the standardized MedMNIST format of 3 x 28 x 28 pixels. This resizing maintains the dataset's utility while ensuring compatibility with the broader MedMNIST framework.

\textbf{Task Description.} The primary application of the BloodMNIST dataset is in various biological and medical image analysis tasks, prominently including the classification of blood cell types. These tasks are crucial for understanding the characteristics of normal blood cells, which can have implications for diagnosing and studying blood-related diseases. The dataset's structure and diverse cell types make it an excellent resource for training and testing machine learning models, especially for classification tasks. Utilizing BloodMNIST provides a means to assess the performance and generalization capabilities of pretrained models in biological imaging, offering insights into how well these models can adapt and apply their learned knowledge to a range of biological tasks.

\subsection*{NF-kB Nuclear Translocation Assay}

\textbf{Dataset Description.} The NF-kB Nuclear Translocation Assay in HCC1143 Cancer Cells dataset focuses on the translocation of the NF-kB protein from the cytoplasm to the nucleus within HCC1143 cancer cells. This process is central to activating the NF-kB pathway, a critical element in cellular responses to various stimuli, especially in cancer biology. The assay encompasses several stages: cell culture, cell seeding, treatment, fixation and permeabilization, and staining. During treatment, cells are exposed to TNF-alpha to induce NF-kB translocation. Staining is performed using specific antibodies for NF-kB and DAPI for nuclear labeling, ensuring clear differentiation between cytoplasmic and nuclear regions. The images acquired through high-content screening fluorescence microscopy provide detailed visualizations of NF-kB translocation post-treatment.

\begin{figure}[h!]
\vskip 0.1in
    \includegraphics[width=\columnwidth]{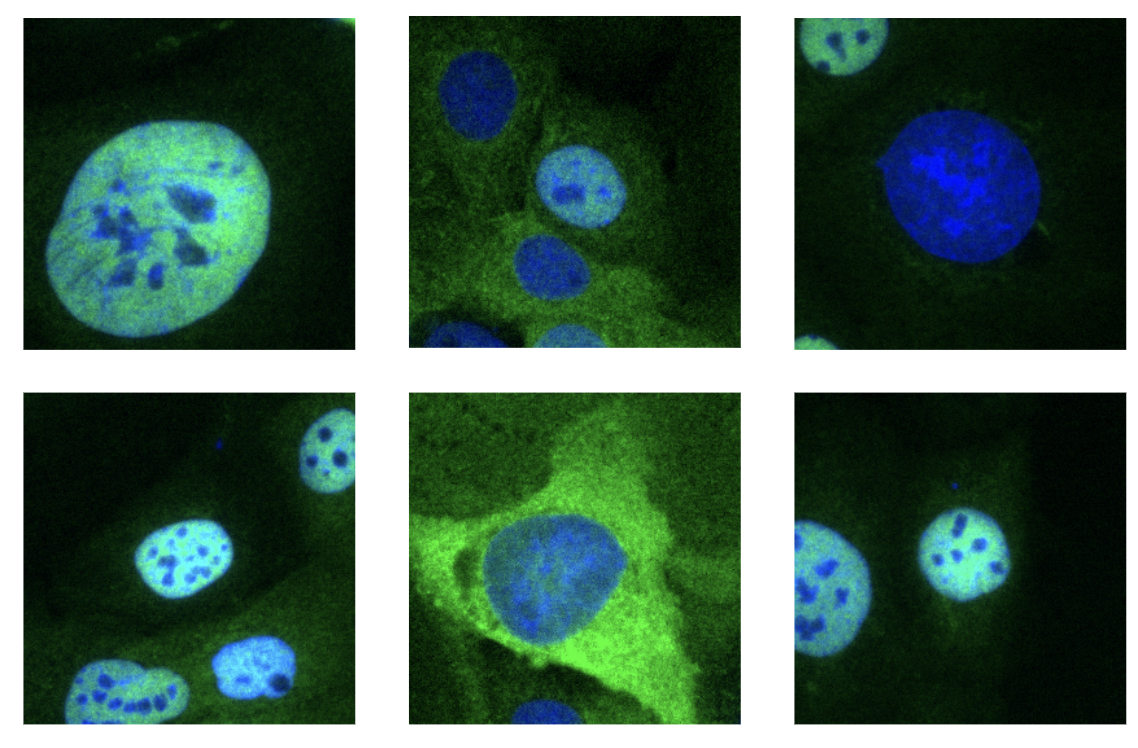}
    \centering
    \caption{Representative Images from the NF-kB Nuclear Translocation Assay: Visualizing NF-kB protein movement in HCC1143 cancer cells, post TNF-alpha treatment.}
    \label{fig:transloc}
    \vskip -0.1in
\end{figure}

\textbf{Task Description.} This assay focuses on quantifying NF-kB translocation from the cytoplasm to the nucleus—a crucial regression task in understanding cancer biology. It provides a standardized approach to assess NF-kB activation in response to TNF-alpha. By comparing treated and untreated cells, the assay enables the delineation of NF-kB activation's degree and kinetics. This is pivotal for understanding the dynamics of NF-kB movement and its implications in cancer biology. The data derived from this assay can be used across different cell lines and under varying conditions, making it a valuable tool for future studies aimed at quantifying translocation extent and deciphering its broader implications in the field of cancer biology.

\subsection*{BBBC021}

\textbf{Dataset Description.} The BBBC021 dataset\footnote{Further information is available at \href{https://bbbc.broadinstitute.org/BBBC021/}{BBBC021}.} from the Broad Bioimage Benchmark Collection contains 39,600 images of MCF-7 breast cancer cells. These cells, a relevant model for p53-wildtype breast cancer research, were treated with 113 small molecules at eight different concentrations. The imaging process employed fluorescent microscopy to highlight DNA, F-actin, and Beta-tubulin. This dataset offers an extensive view into the cellular morphology of MCF-7 cells under various pharmacological conditions, making it an invaluable resource for drug discovery and cell biology research.

\textbf{MoA Task Description.} The primary task associated with the BBBC021 dataset is the prediction of drug mechanisms of action (MoA) through image-based phenotypic profiling. This involves analyzing cellular morphological changes induced by a diverse range of chemical compounds. The dataset facilitates the identification of 12 distinct primary mechanisms, providing critical insights into how different compounds affect cellular morphology and behavior. This information is essential in drug discovery, as it helps in understanding the therapeutic effects and MoA of various compounds. The detailed cellular response analysis to these compounds is key to identifying new drug candidates and comprehending the cellular impact of existing drugs, thereby contributing significantly to cancer research and treatment.

\textbf{Channel Prediction Task Description.} In the realm of cell morphology analysis, the ability to predict cellular components imaged in different channels is pivotal. Given the practical limitations in the number of channels that can be imaged before cell degradation, predicting one channel based on others could enhance cell understanding at reduced costs and augment existing datasets. For the BBBC021 dataset, a key task is predicting the Actin component (second channel) using the other two channels. We specifically approach this task in this paper by utilizing only the CLS token representation from the input channels for Actin channel prediction, allowing an assessment of the specific effects and fidelity of our model representations in channel prediction.

\subsection*{Bray et al}

\textbf{Dataset Description.} The Bray et al. dataset\cite{bray_data}, detailed in Gigascience, presents an extensive array of images and morphological profiles derived from the Cell Painting assay. It encompasses 919,265 five-channel fields of view, featuring images from 30,000 small-molecule treatments and covering 30,616 distinct compounds. This dataset is particularly valuable for its detailed morphological features extracted from individual cells and at population levels, including quality-control metrics and chemical annotations for the compounds used. It stands as a significant resource for comparing cellular states under various chemical perturbations, offering a comprehensive view of cell morphology and function under diverse treatment conditions.

\textbf{MoA Task Description.} The primary task associated with the Bray et al. dataset is, similarly to BBBC021, to explore Mechanisms of Action (MoA) and cellular responses to a wide range of chemical treatments. This involves analyzing how different small molecules impact cell morphology and function. The dataset's extensive collection of images and detailed morphological profiles allow for in-depth studies of the effects of these treatments at both the single-cell and population levels. Such analyses are crucial in cellular biology research, providing insights into the diverse impacts of small molecules on cells. This dataset serves as a rich foundation for developing and testing computational methods in drug discovery and cell biology, aiding in the advancement of research in these fields.

\section{Model details}

\subsection*{Architectural Differences}

\begin{figure*}[tb]
\vskip 0.1in
\begin{center}
\centerline{\includegraphics[width=\textwidth]{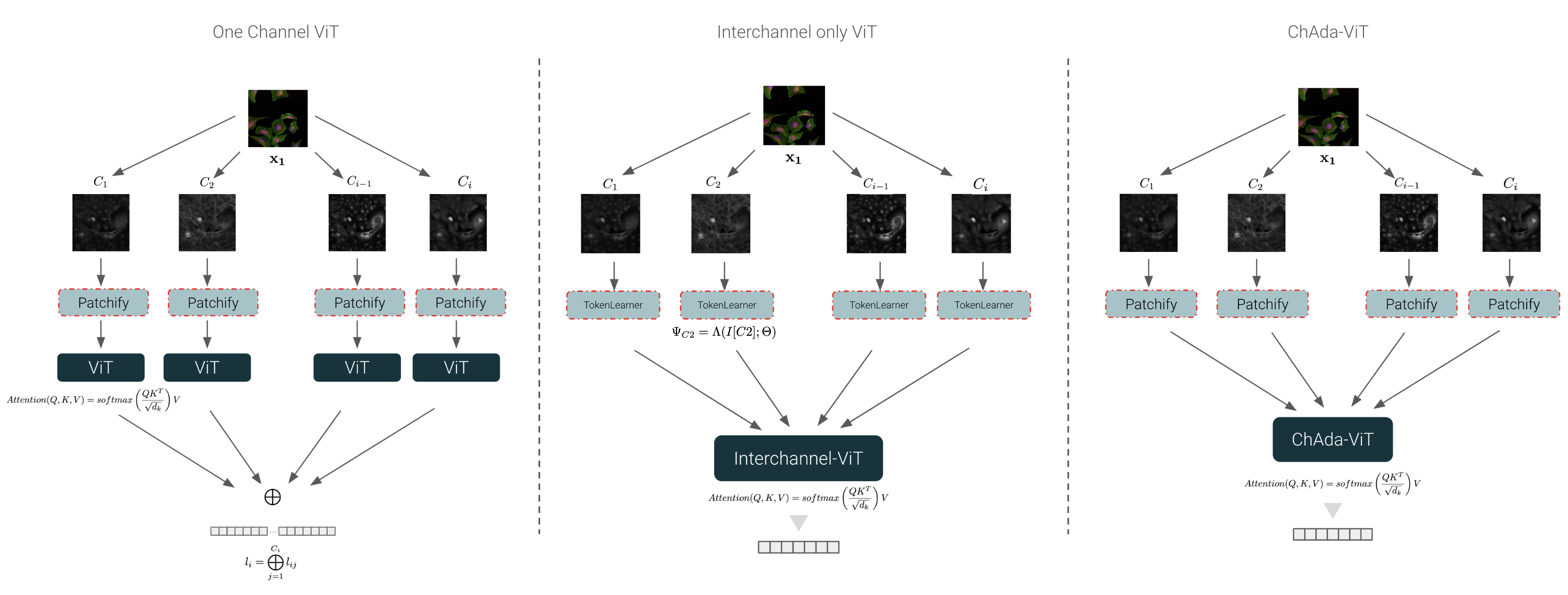}}
    \caption{Comparative Architectural Overview: This figure illustrates the distinct methodologies of One Channel Encoding (left), Inter-Channel Only (center) and ChAda-Vit (right) approaches in bioimage processing. It highlights the unique processing strategies and attention mechanisms employed by each method to handle multi-channel bioimages.}
    \label{fig:archi_diffs}
\end{center}
\vskip -0.1in
\end{figure*}

In the field of bioimaging, different approaches are adopted to encode images with multiple channels into a unified representation, each with its distinct methodology and outcomes, as shown in Figure \ref{fig:archi_diffs}.

The One Channel Encoding approach, prevalent in existing works, treats each channel of an image independently, creating a latent representation for each channel which are then concatenated. This method, while effective for images with uniform channel counts, faces challenges with heterogeneous datasets. It lacks the capacity to encode various datasets into a single representation, limiting its applicability for diverse bioimage datasets.

ChAda-ViT, or Channel Adaptive Vision Transformer, presents a novel approach. It encodes images with varying channel dimensions into a single fixed-size embedding space, incorporating both Inter-Channel and Intra-Channel attention. This method leverages token padding and masking techniques, adapted from NLP transformers, to handle images with different numbers of channels. Additionally, it introduces channel embeddings to preserve channel-specific information. ChAda-ViT’s architecture enables it to process a greater number of input tokens compared to standard Vision Transformer models, thus accommodating the variability in channel count while maintaining the integrity of the self-attention mechanism.

Another variant in this realm is the Inter-Channel only approach. This method tokenizes each channel as a single large patch, compelling the model to focus solely on inter-channel relationships. By omitting Intra-Channel attention, this approach concentrates on the features derived from the relationships between individual channels, differing from the one-channel approach and ChAda-ViT, which consider both intra and inter-channel dynamics.

\subsubsection*{Patch-wise Channel Processing}
Given an image \( I \) of dimensions \( H \times W \times C \), where \( H, W, \) and \( C \) denote the height, width, and number of channels respectively, we dissect each channel into \textbf{\textit{non-overlapping}} patches. For each channel \( c \), the patch at spatial location \( (i, j) \) is denoted as \( P_{c, i, j} \), and is of dimensions \( p \times p \) where \( p = 16 \) in our optimal model configuration.

\noindent Formally, the process of patch extraction and linear projection can be expressed as follows:

\[
P_{c, i, j} = I[c, i \cdot p : (i+1) \cdot p, j \cdot p : (j+1) \cdot p]
\]

\noindent Subsequently, each patch \( P_{c, i, j} \) is linearly projected using a shared Conv2D layer \( f \) with learned parameters \( \theta \), yielding the projected patch \( \hat{P}_{c, i, j} \):

\[
\hat{P}_{c, i, j} = f(P_{c, i, j}; \theta)
\]

Thus, \textit{\textbf{each channel of image \( I \) is transformed into a set of linearly projected patches}}, and the operation is performed identically across all channels to maintain a consistent projection space. This results in a set of projected patches \( \hat{P} \) for the entire image \( I \), which are then further processed through subsequent stages of the Channel-Adaptive Vision Transformer architecture.

\subsubsection*{Positional Embeddings}
Positional information is crucial for retaining the spatial layout of an image through the transformation process. To encode this information, we introduce a differentiable positional embedding to each patch. The positional embedding for a patch at spatial location \( (i, j) \) is denoted as \( \text{pos}_{i,j} \).

\noindent Formally, the positional embedding is added to the linearly projected patch \( \hat{P}_{c, i, j} \) from the previous stage as follows:

\[
\tilde{P}_{c, i, j} = \hat{P}_{c, i, j} + \text{pos}_{i,j}
\]

\noindent This operation is performed for each patch, across all channels, ensuring that patches located at the same spatial coordinates, albeit in different channels, receive the same positional embedding. \\

Through this mechanism, the spatial coherence of the original image is preserved across its transformation into the joint embedding space.

\noindent The positional embeddings \( \text{pos}_{i,j} \) are learnable parameters that are optimized during the training process to better capture the spatial relationships among patches.

\subsubsection*{Channel Embeddings}

Channel embeddings are also introduced to encapsulate the channel order information, ensuring the model can discern patches from different channels even when they are at the same spatial location. Let \( \text{chan}_c \) denote the channel embedding for channel \( c \).

\noindent The channel embedding is added to the previously obtained representation \( \tilde{P}_{c, i, j} \) from the positional embedding stage as follows:

\[
\bar{P}_{c, i, j} = \tilde{P}_{c, i, j} + \text{chan}_c
\]

\noindent This operation augments the patch representation with channel-specific information, ensuring that the model can differentiate patches from distinct channels. The channel embeddings \( \text{chan}_c \) are learnable parameters that are optimized during the training process, allowing the model to learn and represent channel-wise order information effectively.

\noindent Through the integration of channel embeddings, our model can robustly handle the differentiation between patches across channels, addressing a critical challenge in processing multichannel images.

\subsubsection*{Padding Mechanism for Uniform Sequence Generation}

In order to standardize the sequence length across different multichannel images, a padding strategy is employed. Let \( C_{\text{max}} \) denote the maximum number of channels across the dataset, and \( C \) denote the number of channels in a given image \( I \). The difference \( D = C_{\text{max}} - C \) denotes the number of missing channels that need to be padded.

\noindent For each missing channel \( d \) where \( d = 1, 2, \ldots, D \), a padding token \( \text{pad}_d \) is generated and \textbf{\textit{appended}} to the sequence of patches. Formally, the padded sequence of patches for image \( I \) is denoted as \( \text{Seq}_{\text{pad}}(I) \) and is given by:

\[ 
\text{Seq}_{\text{pad}}(I) = \{\bar{P}_{c, i, j} \}_{c=1}^C \oplus \{\text{pad}_d\}_{d=1}^D
\]

, where \( \{\bar{P}_{c, i, j} \}_{c=1}^C \) denotes the sequence of channel-augmented patches from the previous stage, \( \{\text{pad}_d\}_{d=1}^D \) denotes the set of padding tokens for the missing channels and where \( \oplus \) denotes the concatenation operation. \\

\noindent This padding mechanism ensures that the sequence length is \textbf{\textit{uniform}} across all images, facilitating a consistent input structure for the subsequent processing within the Transformer architecture.

\subsubsection*{Handling Padded Sequences in Self-Attention}
The self-attention mechanism is central to the Transformer architecture. However, the presence of padded tokens can distort the attention computation. To tackle this, we utilize a \texttt{src\_key\_padding\_mask}
to indicate the locations of the padded tokens, ensuring they are excluded from the attention computation. \\

\noindent Let \( \text{Seq}_{\text{pad}}(I) \) denote the padded sequence of patches for image \( I \) from the previous stage. The \texttt{src\_key\_padding\_mask} is a binary mask of dimensions \( N \times T \), where \( N \) is the batch size and \( T \) is the sequence length, with ones indicating the locations of padded tokens and zeros elsewhere. \\

\noindent This \texttt{src\_key\_padding\_mask} for image \( I \) is constructed as follows:

\[
\text{mask}_{n,t} = 
\begin{cases} 
0 & \text{if } \text{Seq}_{\text{pad}}(I)[t] \text{ is a padded token} \\
1 & \text{otherwise}
\end{cases}
\]

\noindent The self-attention computation is then modified to exclude the influence of padded tokens. Let \( Q, K, V \) denote the query, key, and value matrices respectively, and \( \text{Attn}(Q, K, V) \) denote the original attention computation. The modified attention computation \( \text{Attn}_{\text{mask}}(Q, K, V) \) is given by:

\[
\text{Attn}_{\text{mask}}(Q, K, V) = \text{Attn}(Q \odot \text{mask}, K \odot \text{mask}, V \odot \text{mask})
\]

, where \( \odot \) denotes element-wise multiplication. Through this modification, the self-attention mechanism effectively ignores the padded tokens, ensuring accurate attention computation across the real patches.

\subsubsection*{Class Token Integration}
A distinct class token, denoted as \([CLS]\), is \textbf{\textit{prepended}} to the sequence of patches to obtain a global representation of the image. The class token is a \textit{differentiable} parameter that is optimized during the training process. Let \( \text{Seq}_{\text{pad}}(I) \) denote the padded sequence of patches for image \( I \) from the previous stage. \\

Thus the sequence with the class token, denoted as \( \text{Seq}_{\text{CLS}}(I) \), is constructed as follows:

\[
\text{Seq}_{\text{CLS}}(I) = [CLS] \oplus \text{Seq}_{\text{pad}}(I)
\]

, where \( \oplus \) denotes the concatenation operation. \\

\noindent This updated sequence \( \text{Seq}_{\text{CLS}}(I) \) is then fed into the Transformer, which processes the sequence through its multiple layers of self-attention and feed-forward networks. The final representation of the [CLS] token captures a global representation of the image, which is utilized for downstream tasks. \\ 

\subsubsection*{TokenLearner in Interchannel only ViT}

The TokenLearner mechanism plays a pivotal role in the interchannel-only Vision Transformer architecture. It is essentially a series of convolutional layers with the primary objective of the TokenLearner to segment each channel of an image into one condensed token. This segmentation is akin to creating a mosaic, where each token represents a concentrated summary of information from a specific channel of the image. 

\noindent Formally, the TokenLearner operates as follows:

\[
\Psi_{c} = \Lambda(I[c]; \Theta)
\]

\noindent Here, \( \Psi_{c} \) represents the transformation of channel \( c \) into a tokenized output. The function \( \Lambda \) denotes the convolutional layers within the TokenLearner, characterized by learnable parameters \( \Theta \). The input \( I[c] \) represents the channel \( c \) of the image \( I \).

\noindent The convolutional layers within the TokenLearner are meticulously designed to process each channel \( c \) of the image \( I \), converting them into a series of large tokens. These tokens are the essence of the image's information condensed into more manageable and informative segments.

\noindent This approach allows the TokenLearner to maintain a consistent tokenization process across all channels of each image, ensuring uniformity and coherence in the representation of the image's information. The resultant tokenized channels are then seamlessly integrated into the ViT architecture, enabling the model to process and interpret the image data with enhanced efficiency and precision.

Through this method, the TokenLearner was meant to learn meaningful feature maps for any channel in order to "tokenize" them into a smaller embedding space.

\section{Experimental Evaluation}

Our experimental approach encompasses several distinct methods to assess the performance of our models across a range of tasks. All experiments were performed under 5 different seeds, and the mean and standard deviation of the results were reported :

\textbf{Linear Probing:} For datasets BloodMNIST, BBBC048, CYCLoPs, and NF-kB Nuclear Translocation Assay (Transloc), we employed linear probing. This involved freezing the encoder of our model and appending a trainable linear layer to the CLS token. The linear layer was then trained specifically for the task associated with each dataset. This method allows us to evaluate the representational quality of the encoded features in a variety of biological contexts.

\textbf{Channel Reconstruction:} In this task, we focus on reconstructing a target channel from given input channels. The encoder is kept frozen, and a simple decoder is added atop the CLS token. The decoder comprises of two fully connected layers, five convolutional layers, followed by a sigmoid function, to predict the target channel accurately. We ensured that the decoder was scaled to maintain an equivalent number of parameters (approximately 5.2 million) for both the One Channel Approach and ChAda-ViT, facilitating a fair comparison of their reconstruction capabilities.

\textbf{Performance Metrics:} For classification tasks, we utilize top-1 accuracy as our primary metric. For regression tasks, the R2 score is employed to measure the accuracy of our predictions. Specifically, for the Channel Prediction task, we use Mean Absolute Error (MAE) and Mean Squared Error (MSE), along with the R2 score. Here, the prediction involves comparing the flattened predicted channel against the flattened target channel. This comparison is conducted on the entire dataset in a single evaluation, providing a comprehensive view of our model's prediction accuracy.

\section{Additional Evaluations}

\begin{figure*}[tb]
\vskip 0.1in
\begin{center}
\centerline{\includegraphics[width=\textwidth]{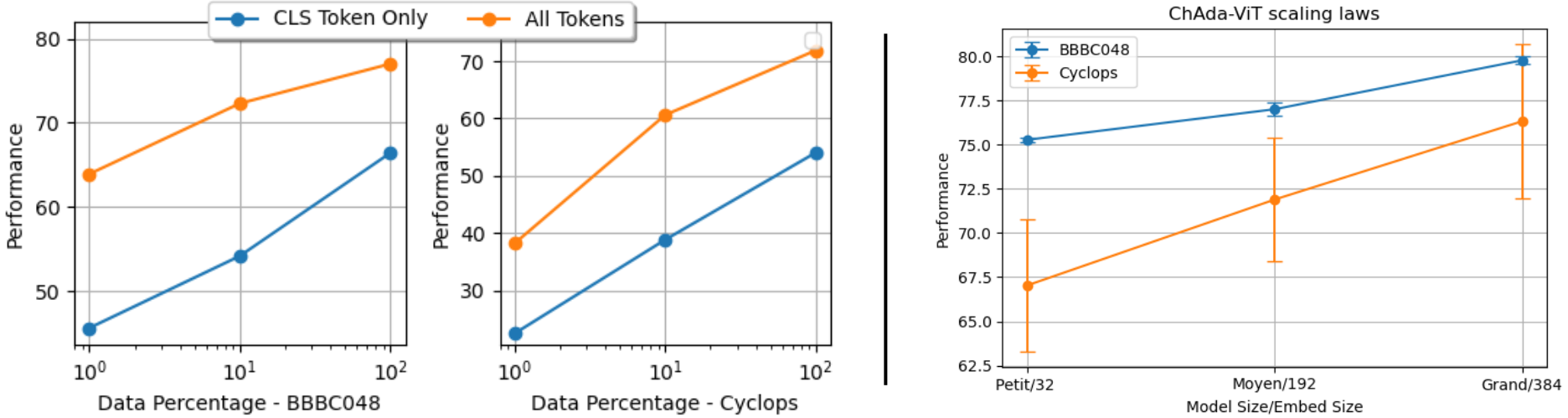}}
\caption{\textbf{Left :} Comparison of evaluation results using ChAda-ViT Moyen/16 with CLS token eval and All output token eval. All token evaluation shows a higher performance than using CLS token. \textbf{Right :} Scaling laws of the performance of the model with All Token Eval, when trained with more parameters and larger embedding sizes, confirming the scalability of the ChAda-ViT architecture.}
    \label{fig:all_token_eval_scaling_laws}
\end{center}
\end{figure*}

\

\begin{table}[h]
\resizebox{\columnwidth}{!}{%
\begin{tabular}{l|rr|rr|}
\cline{2-5}
\textbf{}                                     & \multicolumn{2}{l|}{Linear Eval}                            & \multicolumn{2}{l|}{KNN Eval}                                    \\ \cline{2-5} 
                                              & \multicolumn{1}{l|}{BBBC048} & \multicolumn{1}{l|}{Cyclops} & \multicolumn{1}{l|}{BBBC048} & \multicolumn{1}{l|}{Cyclops} \\ \hline
\multicolumn{1}{|l|}{ChAda-ViT Moyen16}       & \multicolumn{1}{r|}{$77\pm0.38$}      & $71.89\pm3.49$                        & \multicolumn{1}{r|}{$78.37\pm0.22$}   & $51.12\pm3.87$                        \\ \hline
\multicolumn{1}{|l|}{One Channel ViT Tiny/16} & \multicolumn{1}{r|}{$77.48\pm0.14$}   & $70.9\pm3.02$                         & \multicolumn{1}{r|}{$78.43\pm0.04$}   & $38.82\pm0.08$                        \\ \hline
\multicolumn{1}{|l|}{3 Ch. ViT Tiny/16}       & \multicolumn{1}{r|}{$72.08\pm0.14$}   & $40.2\pm2.36$                         & \multicolumn{1}{r|}{$77.79\pm0.02$}   & $37.24\pm2.38$                        \\ \hline
\end{tabular}
}
\caption{Using All Token Evaluation setting, ChAda outperforms both One Channel Approach and the standard 3 Channel ViT (trained on the 3 channel subset of the IDRCell100k), proving the added value of integrating channel level attention over classic approaches.}
\label{tab:all_token_complete_eval}
\end{table}

\begin{figure*}[h]
\vskip 0.1in
\begin{center}
\centerline{\includegraphics[width=\textwidth]{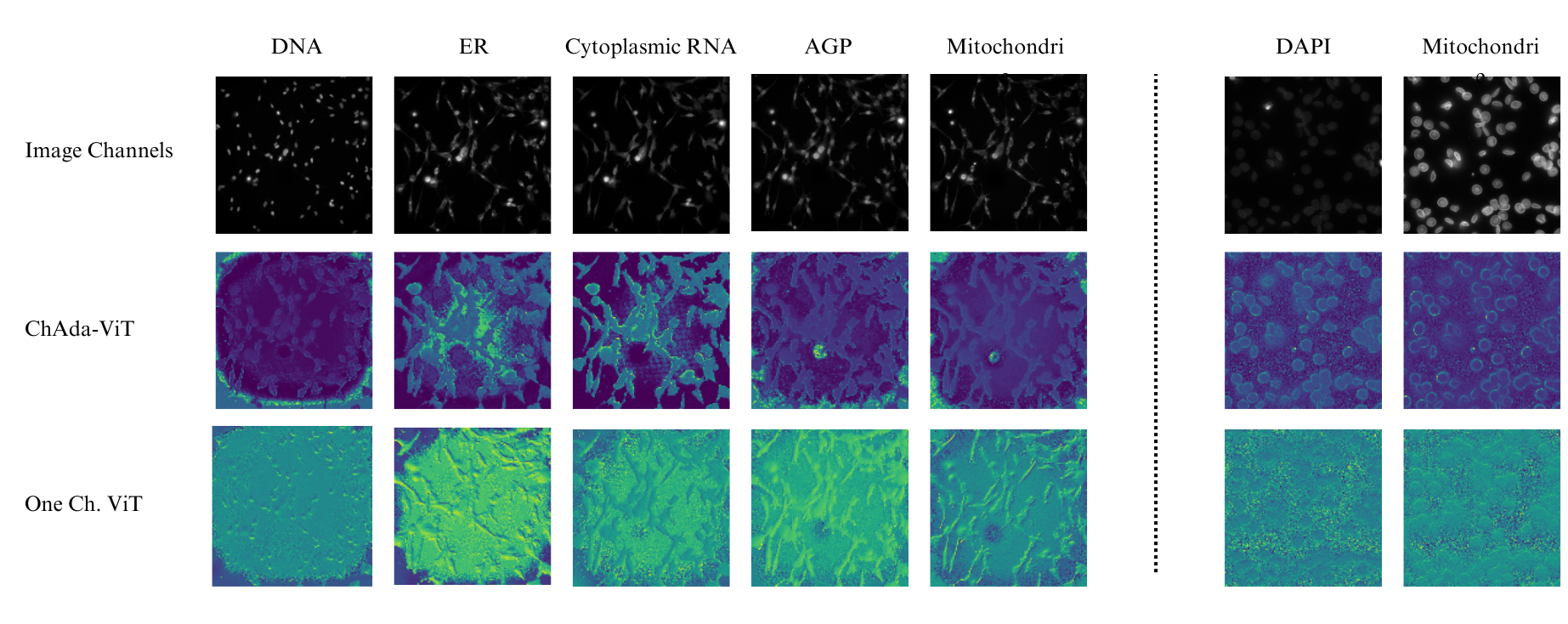}}
\caption{Additional Attention Maps for images with 2 channels (right) and 5 channels (left), with One Channel ViT and ChAda-ViT.}
    \label{fig:additional_maps}
\end{center}
\vskip -0.1in
\end{figure*}

\begin{figure*}[h]
\vskip 0.1in
\begin{center}
\centerline{\includegraphics[width=\textwidth]{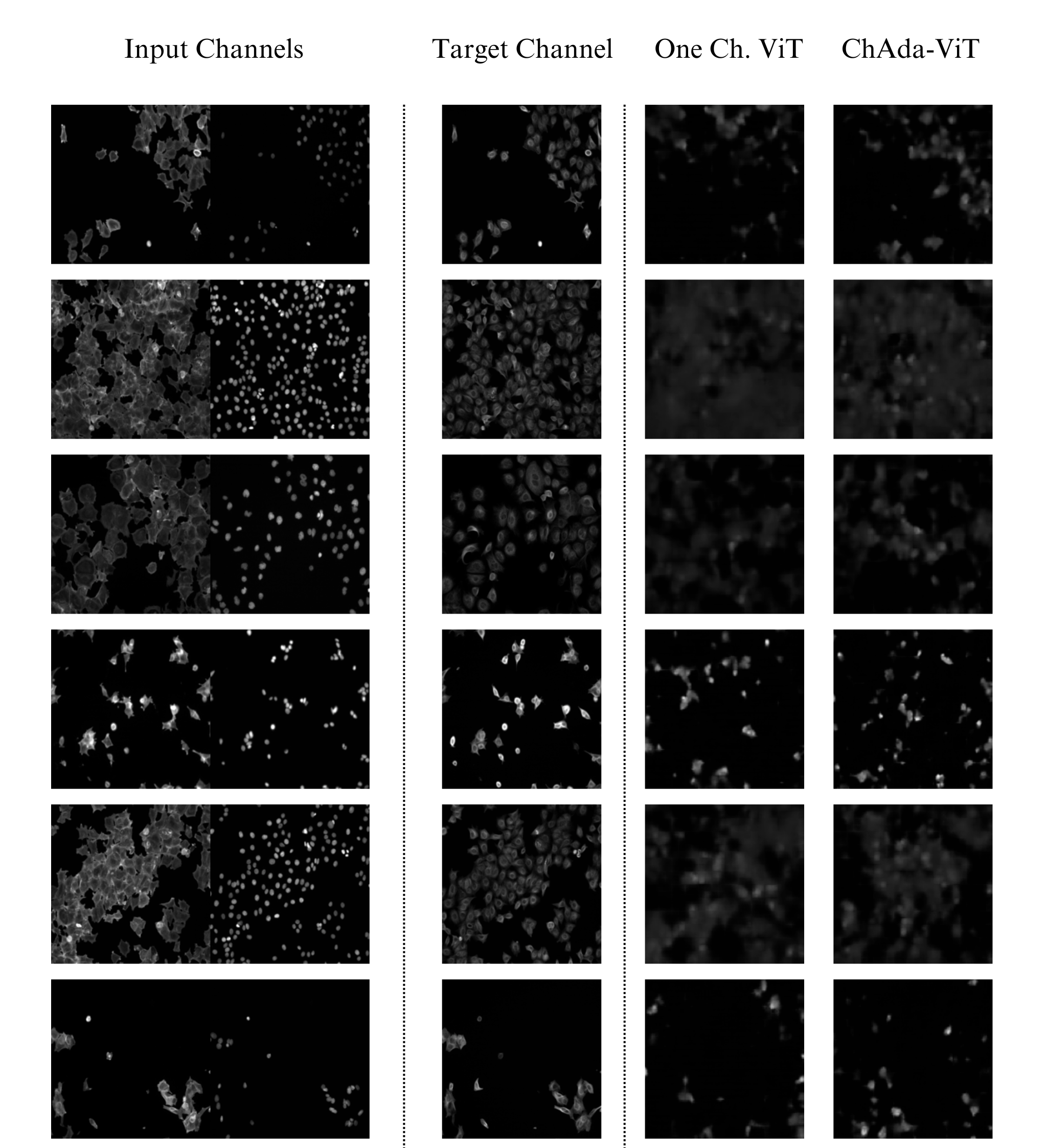}}
\caption{Channel Prediction in BBBC021: Demonstrating ChAda-ViT's enhanced spatial distribution accuracy in reconstructing cell images, compared to the One Channel approach. This superiority is evident even with a basic convolutional decoder utilizing only the \textit{CLS} token, as highlighted in the first row of examples.}
    \label{fig:channel_reconstruction}
\end{center}
\vskip -0.1in
\end{figure*}

\end{document}